\RequirePackage{fix-cm}
\documentclass[smallextended]{for_arxiv}    
\smartqed  
\usepackage{hyperref}
\usepackage{graphicx}
\usepackage{mathptmx}      
\usepackage{hyperref}
\usepackage{amsfonts,amsmath,amssymb}
\usepackage{bbm}

\usepackage[final]{review} 

\begin{document}

\title{Robust Temporally Coherent Laplacian Protrusion Segmentation of 3D Articulated Bodies
}

\titlerunning{Temporal Laplacian Protrusion Segmentation}        

\author{Fabio Cuzzolin \and Diana Mateus \and Radu Horaud
}

\authorrunning{F. Cuzzolin, D. Mateus \& R. Horaud} 

\institute{F. Cuzzolin \at
              Department of Computing and Communication Technologies, Oxford Brookes University, Oxford, UK \\
              \email{fabio.cuzzolin@brookes.ac.uk}           
           \and
           D. Mateus \at
              Institut f\"ur Informatik,
Technische Universit\"at M\"unchen,
Garching b. M\"unchen, Germany\\
\email{mateus@in.tum.de} 
              \and
              R. Horaud \at
              INRIA Grenoble Rh\^one-Alpes, 655 avenue de l'Europe, Montbonnot Saint-Martin, France\\
              \email{radu.horaud@inria.fr}
}


\maketitle

\begin{abstract}
In motion analysis and understanding it is important to be able to fit a suitable model or structure to the temporal series of observed data, in order to describe motion patterns in a compact way, and to discriminate between them. In an unsupervised context, i.e., no prior model of the moving object(s) is available, such a structure has to be learned from the data in a bottom-up fashion. In recent times, volumetric approaches in which the motion is captured from a number of cameras and a voxel-set representation of the body is built from the camera views, have gained ground due to attractive features such as inherent view-invariance and robustness to occlusions. Automatic, unsupervised segmentation of moving bodies along entire sequences, in a temporally-coherent and robust way, has the potential to provide a means of constructing a bottom-up model of the moving body, and track motion cues that may be later exploited for motion classification. Spectral methods such as locally linear embedding (LLE) can be useful in this context, as they preserve ``protrusions", i.e., high-curvature regions of the 3D volume, of articulated shapes, while improving their separation in a lower dimensional space, making them in this way easier to cluster. In this paper we therefore propose a spectral approach to unsupervised and temporally-coherent body-protrusion segmentation along time sequences. Volumetric shapes are clustered in an embedding space, clusters are propagated in time to ensure coherence, and merged or split to accommodate changes in the body's topology. Experiments on both synthetic and real sequences of dense voxel-set data are shown. This supports the ability of the proposed method to cluster body-parts consistently over time in a totally unsupervised fashion, its robustness to sampling density and shape quality, and its potential for bottom-up model construction.
\end{abstract}

\section{Introduction, Background and Contributions}

In motion analysis and understanding it is crucial to be able to fit a suitable model to a temporal series of data observed in association with an object's evolution, in order to describe motions in a compact way, discriminate between them, or infer the pose, however described, of the moving object. In an unsupervised context in which no prior model of the moving object(s) is available, such a structure has to be learned from the data in a bottom-up fashion. \addnote[capture]{1}{In recent times, multi-view stereo methods, e.g., \cite{Furukawa05siggraph,Hernandez04silhouette}, using a set of synchronized and calibrated cameras and yielding volumetric or surface representations, have gained ground, due to attractive features such as inherent view-invariance and relative robustness to occlusion. Moreover, it is also possible to extract 3D motion flows from this kind of data, e.g., \cite{Pons07multi}.}

Meshes, in particular, are widely used in computer vision, computer graphics and computer-aided design to describe objects for a variety of tasks such as object recognition, deformation, animation, visualization, etc. Often, these tasks require the objects to be decomposed into sub-parts: mesh segmentation, therefore, plays a crucial role in many applications, and is a well-studied problem in computer graphics \cite{Huang09shape,Katz05mesh,Lien07approximate,Shamir08survey,Shapira08consistent}, as it links skeletal representations useful for animation with mesh representations necessary for rendering. 
Most approaches, however, deal with smooth meshes that are acquired from artists or from laser-scans of real actors. In opposition, multiple-stereo methods for 3D reconstruction, while they are true to reality at every instant and allow the texture information to be mapped in a straightforward way to the reconstructed geometry, may generate topological and geometric artifacts when assuming no prior knowledge of the scene being observed \cite{Zaharescu2011topology}. For example, the two legs of a human actor may get clubbed together, or holes can be introduced in the resulting 3D model because of silhouette extraction problems \cite{Franco2009silhouettes}. In addition, the obtained meshes are not usually smooth and suffer from local surface distortions due to image and silhouette noise.
As they often make use of surface metrics, such as curvature, which are extremely sensitive to noise, or they assume genus zero shapes \cite{Katz05mesh}, the most common segmentation approaches in computer graphics are limited in their applicability towards visually acquired meshes. The same remarks hold for volumetric representations of the object(s) of interest as sets of voxels. 

More recently, temporal segmentation of visually reconstructed meshes has been addressed as well. \cite{varanasi} proposes a method that performs both convex segmentation of a static mesh and temporally-coherent segmentation of a mesh sequence. \cite{Franco2011learning} proposes a generative probabilistic method that alternates between segmenting a mesh into rigid parts and estimating the rigid motion of each part along time. These methods perform temporal segmentation in the mesh/voxel domain and do not take advantage of the properties of spectral embeddings of articulated shapes.


Mathematically, both mesh and voxel representations can be thought of as graphs describing either the surface or the volume of a shape, in which each vertex of the graph represents a 3D point, while each graph edge connects vertices representing adjacent 3D points. Therefore, mesh/voxel segmentation  can be addressed in the framework of graph partitioning.
\emph{Spectral graph theory} \cite{Cvetkovic98book,Chung97book} provides an extremely powerful framework allowing to cast the graph partitioning problem into a spectral clustering problem \cite{Belkin03laplacian,Luxburg07tutorial}.
In particular, it has been shown that the Laplacian embedding of a graph is well suited for representing the graph's vertices into an isometric space spanned by a few eigenvectors of the Laplacian matrix.
The problem has been thoroughly investigated. Spectral methods allowing geometric representations of graphs are well established, and have been applied to automatic circuit partitioning for VLSI design \cite{Alpert99spectral}, image segmentation \cite{shi00normalized}, 2D point data \cite{Ng02nips}, document mining \cite{Lafon06pami}, web data \cite{Fouss07random}, and so forth. Standard methods assume that the Laplacian matrix characterizing the data is a perturbed version of an ideal case in which groups of data-point are infinitely separated from each other.
Links between Laplacian eigenmaps and linear locally embedding (LLE) \cite{Belkin03laplacian}, kernel PCA \cite{Bengio04learning}, kernel K-means \cite{Dhillon04kernel}, random walks on graphs \cite{Meila01aistats}, PCA \cite{Saerens04ecml,Fouss07random} and diffusion maps \cite{Lafon06pami} have been since established.

To the best of our knowledge \cite{Liu04segmentation} is the first attempt to apply spectral clustering to meshes using \cite{Ng02nips}. A specific feature of both meshes and voxel sets is, however, that the associated graphs are very sparse and that vertex connectivity is relatively uniform over the graph. Under these circumstances, no obvious partitioning of this kind of graph into strongly connected components that are only weakly interconnected exists: this makes mesh segmentation unfeasible via standard graph-partitioning or spectral-clustering algorithms \cite{Chung97book,Ng02nips}, in which the multiplicity of the zero eigenvalue of the Laplacian corresponds to the number of connected components. Firstly, in the case of meshes there is no ``eigengap" \cite{Luxburg07tutorial}: this makes difficult to estimate the dimension of the spectral embedding, and hence the number of clusters, in a completely unsupervised way. Secondly, the eigenvalues of any large semi-definite positive symmetric matrix can be estimated only approximately; this means that it is not easy to study the eigenvalue multiplicities which play a crucial role in the analysis of the Laplacian eigenvectors \cite{Biyikoglu07book}. Finally, ordering the eigenvectors based on these estimated eigenvalues can be less than reliable \cite{Mateus08cvpr}.
    
In \cite{sharma}, these issues were addressed based on an analysis of the \textit{nodal domains} of a graph \cite{Biyikoglu07book}, and the interpretation of the eigenvectors as the principal components of a mesh \cite{Saerens04ecml}. The theory of nodal domains can be viewed as an extension of the analysis of the Fiedler vector \cite{Chung97book} to the other eigenvectors. 

In order to be applicable to the scenario addressed in this paper, in which time series of measurements are exploited to estimate the pose of a moving object or discriminate between different types of motion, the obtained mesh/voxel segmentation has to be \emph{consistent} throughout the temporal sequence. In particular, we are interested in surfaces or volumes representing articulated objects, i.e., sets of rigid bodies linked by articulations. Several approaches have been proposed which rely on the availability of sets of correspondences between pairs of volumes/surfaces. For instance, geodesic distances on both meshes and volumes are invariant to pose and articulation, and can be used for \emph{matching} surfaces which differ by a non-rigid deformation \cite{Bronstein06generalized}. However, they are very sensitive to topological changes (which, as we recalled, often occur in visually reconstructed meshes). \cite{Bronstein09ijcv} uses a metric based on diffusion over the surface through a heat kernel, which is less sensitive in this regard. Functions defined on the mesh-volume are more resistant to topological changes. \cite{Shapira08consistent} uses the local ``thickness" of the mesh-volume to consistently partition different surfaces over articulations and pose changes. These and other works in the geometry processing community use surface curvature to derive the final segmentation, which is fine for smooth 3D models but is not reliable for visually reconstructed meshes.
An example of these matching based approaches is \cite{Golovinskiy09siggraph} which consistently segments sets of objects such as chairs and airplanes, starting from point-wise correspondences via global ICP registration, without exploiting the temporal information.

Other matching based approaches rely on the use of a graphical model to explicitly match two surfaces differing from a non-rigid deformation \cite{Anguelov04nips,Starck07correspondence}. These point correspondences can be later used to recover the articulated structure of the shape, as proposed in \cite{Anguelov04recovering}. \cite{Chang08automatic} proposes an interesting alternative to the graphical model approach: it computes a putative set of rigid transformations between surface points on two articulated shapes, and treats these transformations (instead of the points themselves) as the labels for the graphical model.

A general limitation of model-based approaches to mesh/voxel segmentation is when dealing with unknown scenes, as in the case of multiple interacting actors. These limitations can be addressed by exploiting the temporal consistency between individual reconstructions. Preliminarily computing sets of correspondences between surface or volume points can greatly help to enforce such consistency. Unfortunately, surface/volume matching is a very hard problem, its results suffering from the presence of surface occlusions, sensitivity to topological noise and disjoint objects.

\begin{figure}[ht!]
\centering
\includegraphics[width=\textwidth]{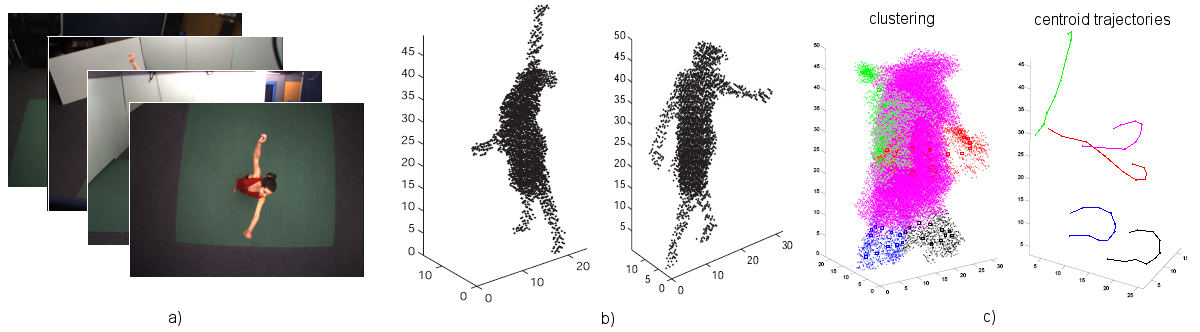}
\caption{Sample results of the proposed approach to unsupervised spectral segmentation of moving bodies (represented as sets of stereo-reconstructed voxels) along entire sequences. Image sequences from multiple cameras (a) are used to generate a volumetric reconstruction of the moving body at each time instant (b); the final result is a consistent unsupervised segmentation of the moving body into segments closely related to the actual body's links, and their centroids' trajectories. \label{fig:contribution}}
\end{figure}

In this paper we focus on volumetric representations, and propose an approach to automatic, unsupervised spectral segmentation of moving bodies (represented as sets of stereo-reconstructed voxels) along entire sequences, in a temporally coherent, robust way, as a means of both inferring a bottom-up model of an unknown articulated body and providing motion cues that can be later used for motion/action recognition, e.g., Figure \ref{fig:contribution} (\addnote[point-cloud]{1}{Throughout the paper, the voxel representations are visualized as point clouds}.) In this endeavor we rely neither on any prior model of the moving body nor on sets of previously computed point correspondences, while accounting for topological artifacts. 

Recent attempts to solve the problem at hand proposed manifold learning techniques to process spatiotemporal data, e.g., \cite{jenkins04icml,lin06a}; these methods rely on enforcing temporal relationships when embedding time sequences, a hard task when handling dense volume representations. We propose instead a mechanism to enforce temporal consistency of segments obtained by clustering shapes in a spectral-embedded space, in which clusters are both collinear and remarkably stable under articulated motions, and propagate them over time. Within the class of spectral embedding techniques that are available today we favor \emph{local linear embedding} (LLE) \cite{Roweis00} for a number of desirable features LLE exhibits in the specific scenario of unsupervised segmentation: while it conserves shape protrusions in virtue of its local isometry, their separation is increased and their dimensionality reduced as an effect of the covariance constraint.

In pose estimation, {\it learning-based} or {\it example-based} approaches \cite{brand99a,grauman03a,elgammal04} directly relate visual information (features) to learned body configurations without the need for accurate a priori models of the studied shape. These methodologies are not affected by the initialization issue, but are limited by the use of training sets of examples. In contrast, techniques have been proposed that directly infer body poses from multiple image cues or volume sequences: when the articulated structure of the moving body is not known at all, the automated recovery of at least a skeleton-like model can be useful prior to any pose estimation step. {\it Skeletonization} methods recover the intrinsic articulated structure of 3D shapes, either directly in 3D \cite{brostow04a}, or in an embedded space \cite{jenkins03cvpr,sundaresan06a}. As it is able to automatically recover the basic structure of an unknown body in terms of protrusions, and to do so consistently along a sequence, the spectral segmentation framework that we propose can be seen as a preliminary step in which both a time-invariant stick-like structure \emph{and} its configuration in the different time instants are recovered.

In \emph{action recognition}, on the other hand, cues need to be extracted from a video sequence in order to recognize the class of motion being performed. While many successful approaches do not rely on dynamics but favor extracting clues from spatiotemporal volumes \cite{action-irani-spacetime-shape-iccv05,kim09pami}, others are based on extracting features, instant-by-instant, and explicitly encode motion dynamics via a graphical model \cite{bissacco07pami,chaudry09histograms,gupta09understanding} such as, for instance, a hidden Markov model \cite{bb69551,bb69593}. In this second case, one has to decide whether to extract features in 2D from individual images, or use multiple-stereo techniques to produce volumetric/surface reconstructions of the moving body. The availability of systems of multiple, synchronized cameras has made this approach both feasible and promising \cite{cuzzolin04icip,cuzzolin04siena}. As the recovered body protrusions are consistently tracked over time along a sequence, the framework proposed here can be easily exploited as a preprocessing stage in which, for example, the 3D locations of the centroids of the recovered segments are used as observations to be fed to a parameter estimation algorithm (such as Baum-Welch for HMMs) in order to identify the performed motion class. 

\addnote[self-citations]{1}{This paper is an extended version of \cite{cuzzolin07workshop} and \cite{cuzzolin08cvpr}. Here we outline the LLE method, which is essential for understanding our temporal-coherent segmentation technique, and we provide an interpretation of LLE in terms of geometric features. A physical interpretation of LLE can be found in \cite{cuzzolin08cvpr}.
The segmentation algorithm, initially proposed in \cite{cuzzolin07workshop} and sketched in \cite{cuzzolin08cvpr}, is described below in more detail. The experimental section contains extensive quantitative evaluations based on synthetic data, which were segmented automatically, as well as more results with real sequences and more comparisons with competing methods. Moreover, with respect to \cite{cuzzolin07workshop,cuzzolin08cvpr} we provide a sensitivity analysis with respect to the LLE free parameters and, in particular, we discuss the algorithm robustness with respect to these parameters, to topological changes in the moving shape, and to the resolution of the voxel grid. We propose an interpretation of protrusion segmentation in terms of the nodal domains of the Laplacian eigenvectors. This in turn provides insights on how to choose the dimension of the embedded space. 
An application to bottom-up model recovery is proposed and illustrated as well.
}

\subsection{Paper Organization}

The remainder of this paper is organized as follows.
Section \ref{sec:problem} starts by outlining the proposed approach, its objectives and the problem constraints. We motivate our approach by the interesting geometric features of LLE. Section \ref{sec:approach} illustrates the proposed algorithm in detail, step by step: the use of \emph{k-wise} clustering to segment the embedded cloud, starting from detected branch terminations (section \ref{sec:branch}), followed by how to propagate clusters over time to ensure consistency (section \ref{sec:propagation}) and merge/split them to fit the topology of the body (\ref{sec:strategy}). The whole procedure is summarized in section \ref{sec:algo}.
Section \ref{sec:exp} describes in detail an extensive set of experimental validations  that provides evidence, on both synthetic and real data, on the performance of the algorithm and in comparison with competing methods. The results are benchmarked with ground-truth data: therefore we thoroughly 
test the way our method and other methods can cope with topology transitions. Moreover we study the robustness of the algorithm. Section \ref{sec:developments} outlines potential applications to bottom-up model recovery. Section \ref{sec:conclusions} concludes the paper.

\section{Problem Formulation and Proposed Solution} \label{sec:problem}

The main objective of this paper is a method for performing unsupervised and temporally coherent protrusion clustering. The proposed methodology should be able to produce an automatic segmentation of an unknown moving body.
Such a body can be represented indifferently as a set of surface points or a set of voxels obtained by volumetric reconstruction from a set of synchronized cameras: here, nevertheless, we will pay special attention to volumetric representations.
Without loss of generality,  the body is supposed to be composed by a collection of linked rigid parts (a condition that could be possibly relaxed).
The produced segments are ``meaningful", in the sense that they are linked to actual segments of the moving body.
The segmentation process takes place in a completely unsupervised way, i.e., neither with previously annotated data nor with human intervention.
The segmentation algorithm processes an entire sequence of surface/volumetric shapes in a consistent way, i.e., the shape is segmented into the ``same" elements throughout the sequence.
The methodology is able to adapt to topological changes due to imperfections or self-contacts.
Throughout the paper, the following constraints are assumed:
\begin{itemize}
\item
no a priori model is available for the moving shape, but instead some model of the latter can be recovered a posteriori, and
\item
no correspondences between sets of surface/volume points are necessary to ensure the consistency of the segmentation (freeing the approach from the limitations of matching algorithms).
\end{itemize}
The remainder of this section explains how spectral segmentation, once LLE is applied, yields geometric features that allow us to design such a system, to be later used for model/pose estimation or action recognition (Section \ref{sec:developments}), and outline an approach to the above defined problem based on it. Then we will detail the various steps of the process, and validate it in a variety of scenarios.

\subsection{Locally Linear Embedding} \label{sec:lle}

\begin{figure}[ht!]
\centering
\includegraphics[width = 0.65\textwidth]{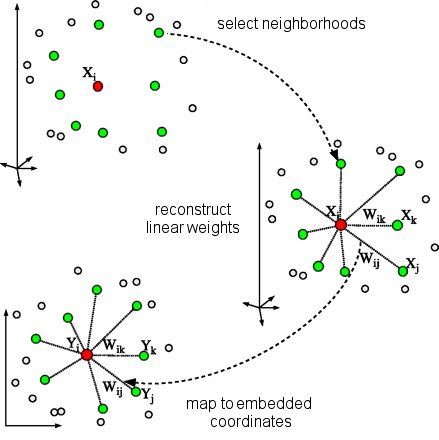}
\caption{Graphical description of the LLE algorithm. \label{fig:lle}}
\end{figure}

LLE \cite{Roweis00} is a graph Laplacian \cite{vgl} algorithm which computes the set of $d$-dimensional embeddings $Y = \{ Y_i, i = 1,..,N \}$ of a set of input points $X = \{X_i, i = 1,..,N\}$, while preserving their local structure. For each data point $X_i$, the algorithm computes the weights $\{ W_{ij}, j=1,...,N(i) \}$ that best linearly reconstruct $X_i$ from its $k$ nearest neighbors $\{ X_j, j = 1,..,k \}$ by solving the following constrained least-square problem: $\arg\min \sum_{i=1}^N | X_i - \sum_{j=1}^{k} W_{ij} X_j |^2$. Then the low-dimensional embedded vectors $\{Y_i\}$ associated with the original cloud of points $\{X_i\}$ are computed by minimizing the sum of the squared reconstruction errors (Figure \ref{fig:lle}):
\begin{equation} \label{eq:opt}
\arg\min_{\{Y_i\}} \sum_{i=1}^N \Big | Y_i - \sum_{j=1}^{k} W_{ij} Y_j \Big |^2.
\end{equation}
The embedded cloud $Y$ is constrained to be centered at the origin $\mathbf{0}$ of $\mathbb{R}^d$ ($\sum_i Y_i = \mathbf{0}$) and to have unit covariance:
\begin{equation} \label{eq:covariance}
\frac{1}{N} \sum_{i=1}^N Y_i Y_i^T = I.
\end{equation}
The objective function to minimize (\ref{eq:opt}) is in fact a quadratic form $\sum_{i,j = 1}^N M_{ij} \langle Y_i, Y_j \rangle$ (where $\langle . \rangle$ denotes the usual scalar product of vectors) involving the ``affinity" matrix:
$M \doteq (I-W)^T (I-W)$, $W = [w_{ij}]$. The optimal embedding (up to a global rotation) is found by computing the bottom $d+1$ eigenvectors of $M$, and discarding the bottom (unit) one.

\subsection{Geometric Features of LLE} \label{sec:lle-features}

\begin{figure}[ht!]
\centering
\includegraphics[width = \textwidth]{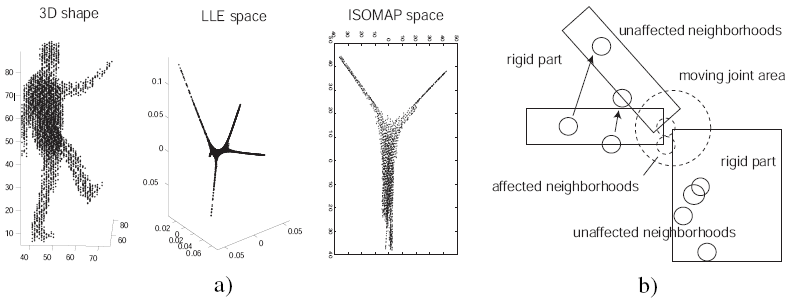}
\caption{a) How LLE (middle) and ISOMAP \cite{Tenenbaum00isomap} (right) map the same 3D cloud (left) for the same number of neighbors $k=13$. LLE increases the separation of protrusions like legs and arms, unlike methods based on geodesic distances. b) The number of neighborhoods affected by articulated motion is relatively small. \label{fig:clustering}}.
\end{figure}

Despite having been originally introduced as an unsupervised dimensionality reduction technique, LLE displays a number of attractive geometric features for unsupervised segmentation as well, as it is illustrated by Figure \ref{fig:clustering}. Given a cloud of 3D points representing, for instance, the voxels occupying the volume of an articulated shape, for a large interval of its parameters $k$ (scale of the neighborhood to preserve) and $d$ (dimensionality of the embedding space), the corresponding embedded cloud after LLE exhibits two interesting properties:
\begin{enumerate}
\item
it is (roughly) \emph{lower dimensional}, i.e, the embedded points live (approximately) on a manifold of dimension lower than the original one (a 1-D tree in Figure \ref{fig:clustering}-a);
\item
the protrusions (chains of smaller rigid links) in the original shape are mapped to roughly equally spaced, widely separated branches of the embedded cloud.
\end{enumerate}
Intuitively, this is due to the fact that the covariance constraint (\ref{eq:covariance}) pulls outwards and stretches the chain of links formed by all local neighborhoods, redistributing them on a (roughly) lower dimensional manifold. As the force pulls those chains in a radial direction, their separation increases in the embedding space. Clustering/segmenting such protrusions becomes much easier there: compare in Figure \ref{fig:clustering}-a)-right the output of the ISOMAP embedding on the same shape. 

\begin{figure}[ht!]
\centering
\includegraphics[width = \textwidth]{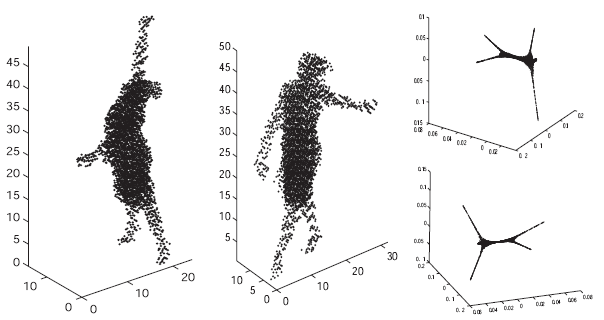}
\caption{Stability of LLE under articulated motion. Two different poses (left and middle) of the same person are mapped onto the same embedded shape (right), for a large interval of parameter values and up to an orthogonal transformation. \label{fig:pose-invariance}
}
\end{figure}

However, our purpose is to cluster moving objects in a temporally coherent way: we need the segments obtained at different time instants to be \emph{consistent}, i.e., to cover the same body-parts. Some embedding schemes, such as ISOMAP, are inherently \emph{pose-invariant} under articulated motion (as, while the articulated body evolves, the geodesic distances between all pairs of points do not vary). This is not true, in a strict sense, for LLE and other graph Laplacian embeddings \cite{Belkin03laplacian}. However, for articulated bodies formed by a number of rigid parts linked by rotational joints, all local neighborhoods incident on a rigid part are preserved along the motion, while only the few neighborhoods interested by the evolving joint(s) are affected (Figure \ref{fig:clustering}-b). Figure \ref{fig:pose-invariance} illustrates two different poses of a dancer performing a ballet, and how the related embedded clouds obtained through LLE for $d=3$ and $k = 10$ are stable (up to an orthogonal transformation).

\subsection{The Proposed Formulation} \label{sec:proposed-approach}

Shape protrusions are then conserved in the LLE spectral domain, while their separation is increased and their intrinsic dimensionality reduced. The effect is dramatic in situations in which body-parts are physically close to each other in the original space, and makes clustering significantly easier after embedding. Moreover, the embedded shape of an articulated body is to a large extent pose-invariant, making the propagation in time of the obtained clusters much easier.

Here we are going to exploit these geometric features of LLE to devise an \emph{unsupervised}, time-consistent method for protrusion segmentation of 3D articulated shapes. The proposed segmentation technique does not assume any prior model (not even a weak topological or graphical one). The number of clusters themselves is unknown while it is inherently determined by the lower dimensional structure of the embedded cloud. The proposed method follows the following steps:
\begin{enumerate}
\item
At each time-instant the current 2D voxel-cloud is mapped via LLE to a suitable embedding space;
\item
The most suitable number of clusters is estimated by counting the number of branches in the embedded cloud;
\item
The $k$-wise clustering technique devised in \cite{bpc} is used to segment such branches, due to their intrinsic lower dimensionality;
\item
Such a embedded-space clustering induces protrusion segmentation in the original 3D space, and finally
\item
The resulting segmentation is propagated to the next time-instant to ensure temporal coherence of the segmented body.
\end{enumerate}
In the next section we are going to provide a detailed description of each step. 

\section{Method}

\label{sec:approach}

As articulated shapes are mapped by LLE onto an embedded space, it would make no sense to employ generic k-means \cite{macqueen67kmeans} to segment embedded-space protrusions (as it happens in classical spectral clustering \cite{Ng02nips}). We therefore adopt the \emph{k-wise clustering} introduced in \cite{bpc} to segment protrusions in the embedding space. Given a sequence of 3D shapes, the initial segmentation is \emph{consistently propagated} along time, and the number of clusters estimated in an automatic way to fit the changing topology of the moving articulated body.

\subsection{K-wise Clustering of the Embedded Shape} \label{sec:kwise}

\begin{figure}[ht!]
\centering
\includegraphics[width = \textwidth]{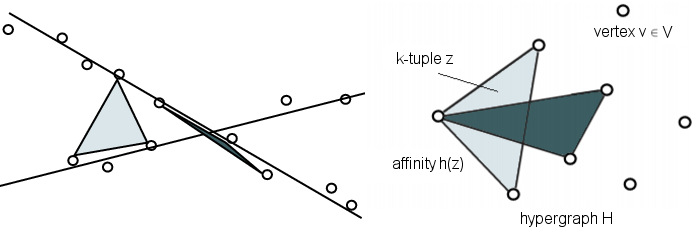}
\caption{Left: rationale of 3-wise (k-lines) clustering. The areas of the triangles defined by triads of points measure their collinearity. As the points tend towards collinearity, the area of the triangle tends to 0. Right: affinity of $k$-tuples in a hypergraph $H$. \label{fig:klines}}
\end{figure}

Consider Figure \ref{fig:clustering}. For a dimension $d = 3$ of the LLE embedding space, the embedded cloud is (for a wide interval of values of $k$) a tree-like one-dimensional curve. When trying to segment its branches, then, it is natural to look for clusters formed by sets of roughly collinear points. Traditional clustering algorithms, such as k-means \cite{macqueen67kmeans}, are based on measuring pairwise distances between data points. In opposition, it is possible to define measures of similarity between \emph{triplets} of points, which signal how close these triplets are to being collinear (Figure \ref{fig:klines}-left) \cite{heiser97triadic,hayashi72}). More generally, the problem of clustering points based on similarity between $k$-tuples of points is called \emph{k-wise clustering}. An interesting approach to k-wise clustering has been proposed in \cite{bpc}, based on the notion of ``hypergraph".

A weighted undirected hyper-graph $H$ is a pair $(V, h)$, where $V$ is the set of vertices of $H$. Subsets of $V$ of size $k$ are called hyper-edges (just as subsets of two elements in a normal graph are potential edges). The function $h$ associates nonnegative weights $h(z)$ with each hyper-edge ($k$-tuple) $z$, and measures the affinity of each hyper-edge (Figure \ref{fig:klines}-right), in analogy with the weight of edges in conventional graphs.

Consider then a set of data-points $Y = \{ Y_i, i = 1,...,N \}$.
\begin{enumerate}
\item
First, an \emph{affinity hyper-graph} $H$ is built by measuring the affinity of all the $k$-tuples of points in $Y$;
\item
Next, a weighted graph $G$ with the same vertices $Y$ approximating the hyper-graph $H$ is constructed by
constrained least square optimization;
\item
Finally, the approximating graph $G$ is partitioned into $n$ parts via a spectral clustering algorithm \cite{shi00normalized,Ng02nips}.
\end{enumerate}
In the problem of interest to us, the hyper-graph to approximate has as vertices the embedded points $\{Y_i\}$. If the embedding is $d$-dimensional, $d$-tuples of embedded points are considered as hyper-edges: in particular, hyper-edges are  triads of points when $d=3$. A natural choice for the affinity of these triads of embedded points is the area of the triangle they form, e.g., Figure \ref{fig:klines}-left, or the volume of the $(d-1)$-dimensional hyper-edge in the general case.

The outcome of the above clustering algorithm is a segmentation of the embedded cloud $Y$, which can be trivially transferred back to the original 3D space using the known correspondences between the 3D points and the embedded-space points. A mechanism for automatically estimating the number $n$ of clusters is the next step in the algorithm.

\subsection{Branch Detection and Number of Clusters} \label{sec:branch}

\begin{figure}[ht!]
\centering
\includegraphics[width = 0.9\textwidth]{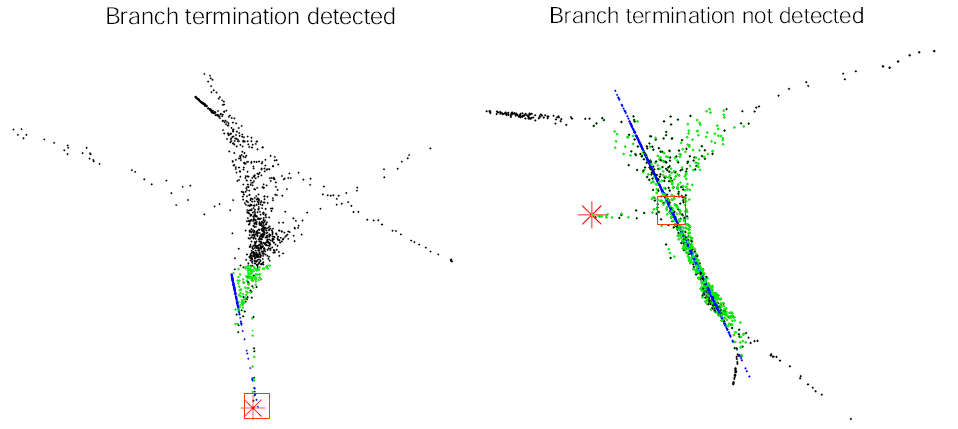}
\caption{Termination (left) and internal (right) points of the
embedded cloud are characterized by their projection (red square) on the line (in blue) interpolating their
neighborhoods (in green) being an extremum of the interval of all projections. \label{fig:termination}}
\end{figure}

The fact that embedded clouds typically appear as one-dimensional strings formed by a number of branches (corresponding to the extremities of the moving body) provides us with a simple method to estimate, at each time
instant, the most suitable number of clusters (Figure \ref{fig:termination}).

Each point of the embedded cloud is tested to decide whether or not it is a branch termination. This test is performed
by looking for its nearest neighbors, for a certain threshold distance empirically learned from the data. The best interpolating line (in blue) for all the neighbors (in green in Figure \ref{fig:termination}) of the considered point is found, and all such neighbors are projected onto it. A point of the embedded cloud (red star) is considered a branch termination if the projections of all its neighbors on the interpolating line lay on one side of its own projection (red square), e.g., Figure~\ref{fig:termination}-left. It is not considered a termination when its projection has neighbors on both sides (Figure \ref{fig:termination}-right).
This algorithm proves to work extremely well on embedded clouds generated through LLE. It becomes then possible to
detect transitions in the topology of the moving body whenever they happen, and modify the number of clusters accordingly.
We will return on this in Section \ref{sec:strategy}.

\subsection{Temporal Consistency and Seed Propagation} \label{sec:propagation}

When considering entire sequences of 3D clouds we need to ensure the \emph{temporal consistency} of the segmentation:
in normal situations (no topology changes due to contact of different body-parts) the cloud has to be decomposed into
the ``same" groups in all instants of the sequence. We propose a propagation scheme in which centroid clusters at time $t$ are used to generate initial seeds for clustering at time $t+1$ (Figure \ref{fig:propagation}). Let $n$ be the number of clusters.\\

\subsubsection{The Seed Propagation Algorithm}

\begin{figure}[t!]
\centering
\includegraphics[width = 0.8\textwidth]{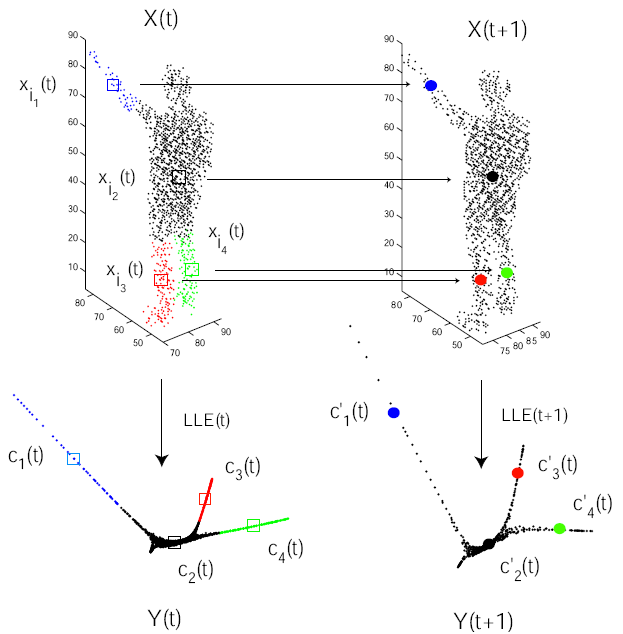}
\caption{Seed propagation for consistent clustering along time in the embedding space. The anti-images of the embedded centroids at time $t$ are added to the 3D cloud $X(t+1)$ at time $t+1$. Their own embeddings $c'_j(t+1)$ are the seeds from which to start clustering the new embedded cloud $Y(t+1)$. \label{fig:propagation}}
\end{figure}

\begin{enumerate}
\item
The embedded cloud $\{Y_i(t), i=1,...,N(t)\}$ at time $t$ is clustered using $d$-wise clustering (Section
\ref{sec:kwise}, Figure \ref{fig:propagation}-bottom-left);
\item
for each cluster centroid $c_j(t)$ ($j = 1,...,n$) in the embedding space (a square in Figure \ref{fig:propagation}-bottom-left), the original datapoint, denoted by $X_{i_j}(t)$, whose embedding $Y_{i_j}(t)$ is the closest neighbor of $c_j(t)$ (and which we call ``3D centroid") is found (a square of the same color in Figure \ref{fig:propagation}-top-left):
\[
i_j(t) = \arg\min_{i = 1,...,N} \| Y_i(t) - c_j(t) \|^2;
\]
\item
at time $t+1$ the dataset of 3D input points $X(t+1) = \{ X_i(t + 1), i=1,..., N(t + 1) \}$ is augmented with the positions of the old ``3D centroids" at time $t$ (colored circles), yielding the set (Figure \ref{fig:propagation}-top-right):
$
X'(t+1) = X(t+1) \cup \{ X_{i_j}(t), j = 1,...,n\};
$
\item LLE is applied to the extended dataset $X'(t+1)$, obtaining (Figure \ref{fig:propagation}-bottom-right) \footnotemark[1]\footnotetext[1]{The embeddings $c'_j(t+1)$ of the previous 3D centroids
$X_{i_j}(t)$ in the new embedded cloud can also be computed by \emph{out of sample extension} \cite{bengio03}.}
$
Y(t+1) \cup \{ c'_j(t+1), j = 1,...,n\},
$
where $c'_j(t+1)$ is the embedded version of the \emph{old} 3D centroid $X_{i_j}(t)$ in the \emph{new} embedded cloud at time $t+1$ (a colored circle in Figure \ref{fig:propagation}-bottom-right);
\item such embedded points $c'_j(t+1)$ are then used as seeds from which to cluster via k-wise the new embedded cloud $Y(t+1)$.
\end{enumerate}

\subsection{Topology Changes and Dynamic Clustering} \label{sec:strategy}

The question of how to initialize the seeds at $t=0$ naturally arises. Besides, even though working in an embedding
space helps to dramatically contain the issue with segmenting body-parts which get dangerously close to each other, instants in which different parts of the articulated body come into contact still have important effects on the shape of the embedded
cloud. In fact, in an unsupervised context in which we do not possess any prior knowledge about the number of rigid parts
which form the body or the way they are arranged, but only the fact that they belong to an articulated shape, there is no reason to distinguish touching body-parts. It appears more sensible to \emph{adapt the number of clusters to the number of actually distinguishable protrusions}.

The branch detection algorithm of Section \ref{sec:branch} provides, on one side, a suitable tool for initializing the embedded clustering machinery, while allowing at the same time the framework to adapt the number and location of the clusters when a topology change occurs.

\subsubsection{The Cluster Merging-Splitting Algorithm}

\begin{enumerate}
\item
At each time instant $t$ all the branch terminations of the embedded cloud $Y(t)$ are detected; if
$t=0$ they are used as seeds for $d$-wise clustering;
\item
otherwise ($t>0$), first standard $k$-means with $k=d$ is performed on $Y(t)$ using branch terminations as seeds,
yielding a rough partition of the embedded cloud into distinct branches;
\item
if two or more propagated seeds $c'_j(t)$ (see Section \ref{sec:propagation}) fall inside the same element of this preliminary partition obtained via k-means, they are replaced by the branch termination of the element of the preliminary partition;
\item
for each element of the preliminary partition of $Y(t)$ which does not contain any propagated seed, a new seed is defined as the related branch termination;
\item
finally, $d$-wise clustering is applied to the resulting set of seeds.
\end{enumerate}
Step~3 takes place when previously separated protrusions get too close (in the original 3D space) to be distinguished: it makes then sense to merge the corresponding clusters. Step~4 represents the opposite event in which a body-part which was previously impossible to distinguish becomes well separated, requiring the introduction of a new cluster. As a result, clusters are allowed to merge and/or split according to topological changes in the moving articulated body. We will validate this approach to topology change management in Section \ref{sec:exp5}.

\subsection{Summary of the Method} \label{sec:algo}

Let us at this point summarize our approach for unsupervised, robust segmentation of parts of moving articulated bodies in a
consistent way along a time sequence, by integrating the separate algorithms of Sections \ref{sec:kwise}, \ref{sec:branch},
\ref{sec:propagation}, \ref{sec:strategy} into a coherent whole.\\ Given the values $d,k$ of the parameters of LLE, for each time instant $t$:
\begin{figure}[ht!]
\centering
\includegraphics[width = 1 \textwidth]{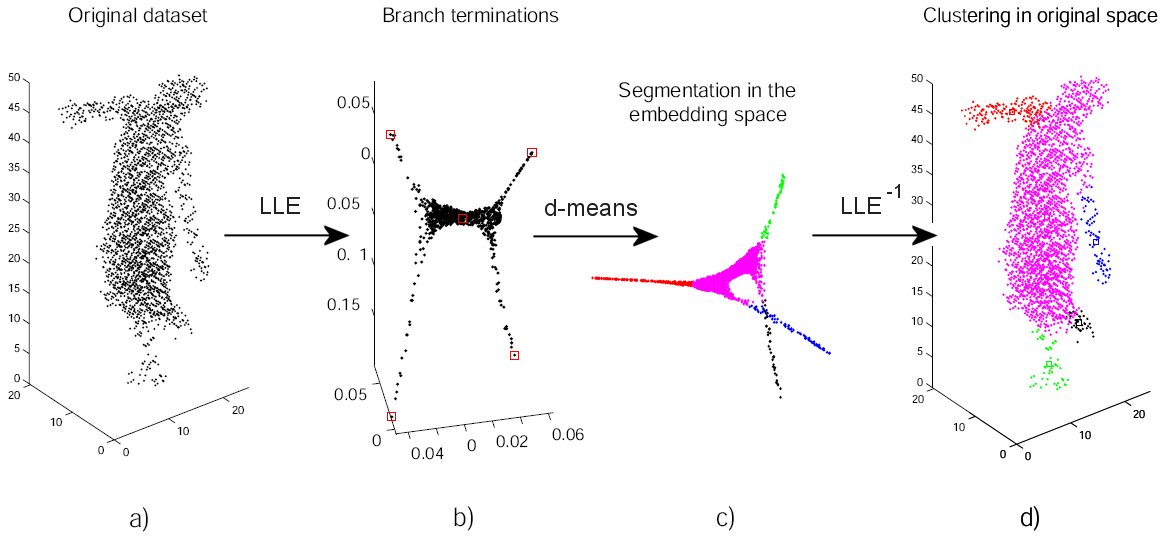}
\caption{Graphical illustration of the proposed coherent spectral segmentation algorithm. \label{fig:algo}}
\end{figure}
\begin{enumerate}
\item
the current data-set $X'(t) = X(t) \cup \{ X_{i_j}(t-1) \}$ for $t>0$, $X'(0) = X(0) = \{ X_i(0), i =1,..., N(0) \}$ for $t=0$ (Figure \ref{fig:algo}-a), is mapped to an embedding space of dimension $d$, yielding an embedded cloud $Y(t) = LLE[k,d](X'(t))$;
\item
all the branch terminations of the embedded cloud $Y'(t)$ are detected (Section \ref{sec:branch}):
the number of clusters $n(t)$ at time $t$ is then set to the number of branches (plus one for the torso),
Figure \ref{fig:algo}-b;
\item
the embedded cloud $Y'(t)$ is clustered into $n(t)$ groups by $d$-wise clustering (Section
\ref{sec:kwise}) starting from $n(t)$ seeds (Figure \ref{fig:algo}-c):
\begin{itemize}
\item if $t=0$, we use as seeds the branch terminations;
\item if $t>0$, seeds are obtained from the old propagated centroids $\{ c'_j(t), j=1,...,n(t-1) \}$ and the new branch terminations via the splitting/merging procedure described in Section \ref{sec:strategy};
\end{itemize}
\item
this yields a new set of centroids $\{ c_j(t), j=1,..,n(t) \}$ in the new embedding space $Y'(t)$;
\item
the labeling of the embedded points induces a segmentation in the original 3D shape (Figure \ref{fig:algo}-d);
\item
the new cluster centroids $c_j(t)$ are re-mapped to 3D (Section \ref{sec:propagation}), the corresponding 3D
centroids $X_{i_j}(t)$ are added to the data-set $X(t+1)$ at time $t+1$ (Figure \ref{fig:propagation});
\item
we go back to step 1, until the last time instant of the sequence is attained.
\end{enumerate}

\section{Experiments} \label{sec:exp}

\subsection{Experimental Setup}

We tested the algorithm of Section \ref{sec:algo} on both synthetic and real data, in order to have both qualitative and quantitative assessments on its performances.

\begin{figure}[ht!]
\centering
\includegraphics[width = \textwidth]{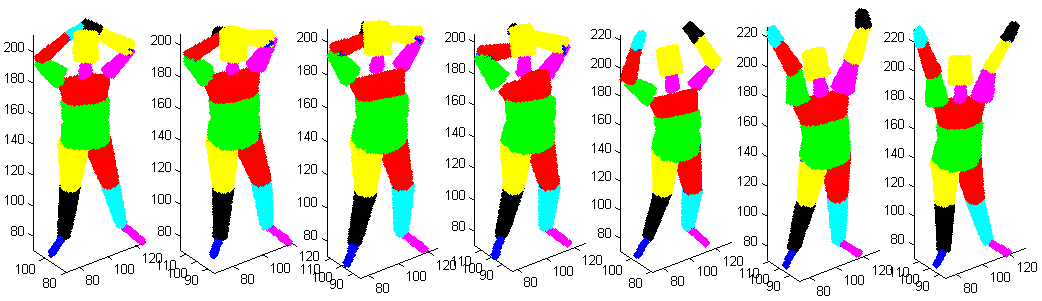}
\caption{A few sample frames from the synthetic sequence ``reveil" used in the first experiment. The automatically generated ground-truth segmentation into rigid links is shown in colors. \label{fig:synthetic} }
\end{figure}

In the \emph{first experiment} (Section \ref{sec:exp1}) we tested our algorithms on a set of synthetic sequences depicting a moving person, for which a ``ground truth" segmentation of the person's body into rigid links was available, e.g., Figure \ref{fig:synthetic}. The sequences were generated by simulating the evolution of a human body model formed by a kinematic model \cite{Knossow2008tracking}, plus the volume elements representing its rigid links. As it can be seen in Figure \ref{fig:synthetic}, the grid sampling was such that the height of a standing person was 200 voxels, corresponding to a real-world resolution of around 1cm.

\begin{figure}[t!]
\centering
\begin{tabular}{c}
\includegraphics[width = \textwidth]{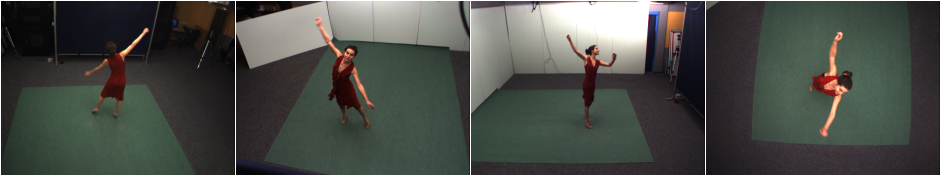} \\ \\ \includegraphics[width = \textwidth]{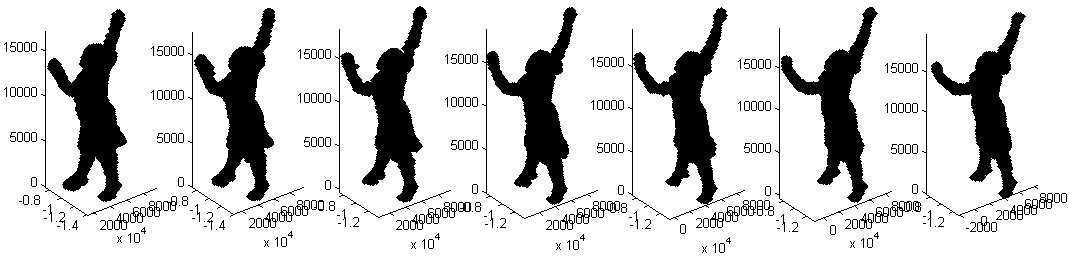}
\end{tabular}
\caption{Top: sample frames of the ``dancer" sequence. Bottom: frames 20-26 of the real voxel-set sequence ``dancer" used in the second experiment. \label{fig:real} }
\end{figure}

In a \emph{second experiment} (Section \ref{sec:exp2}) we used a multiple-camera setup composed by eight synchronized and cross-calibrated video cameras to capture images sequences in a constrained environment, such that the silhouette of a moving person could be easily extracted from each image. Using the resulting silhouettes, a standard space-carving algorithm was employed to generate at each time instant a voxel-based representation of the person. The frame rate of the cameras was 30 frames per second. The original voxel size was 1cm. Figure \ref{fig:real} shows as an example a few frames from the ``dancer", one of the real-world voxel-set sequences, and a number of images for the same sequence\footnote{These data are available on-line at \url{http://4drepository.inrialpes.fr/public/datasets}.}. As it can be appreciated, the original resolution of the voxel-set grid is quite high.

In the case of sequences for which motion capture estimates were not available, and could not be used to provide ground truth on the ``true" segmentation into rigid links, we could nevertheless provide visual results which provide qualitative evidence on the effectiveness of the approach.

In a \emph{third experiment} (Section \ref{sec:exp3}) we were able to calculate performance scores for real-world sequences for which motion capture data was available, and a ground truth segmentation of the moving person could be built. In all these experiments we compared our results with those of similar schemes in which seeds are also passed to the next frame to ensure time consistency, but clustering is either performed in 3D on the original data-set using a Gaussian mixture and the EM algorithm \cite{dempster77em}
(time-consistent EM clustering) or in the ISOMAP space \cite{Tenenbaum00isomap} using k-means (time-consistent ISOMAP clustering).

In Section \ref{sec:exp4} we further analyzed the sensitivity of the algorithm to changes in the values of its basic parameters: the set of eigenvectors selected to compute the desired LLE embedding, the size $k$ of the neighborhood, and the resolution of the grid of voxels. Finally, in Section \ref{sec:exp5} the robustness with respect to topological changes in the moving shape was assessed.

\subsection{Ground Truth and Performance Scores}

\begin{figure}[t!]
\centering
\includegraphics[width = \textwidth]{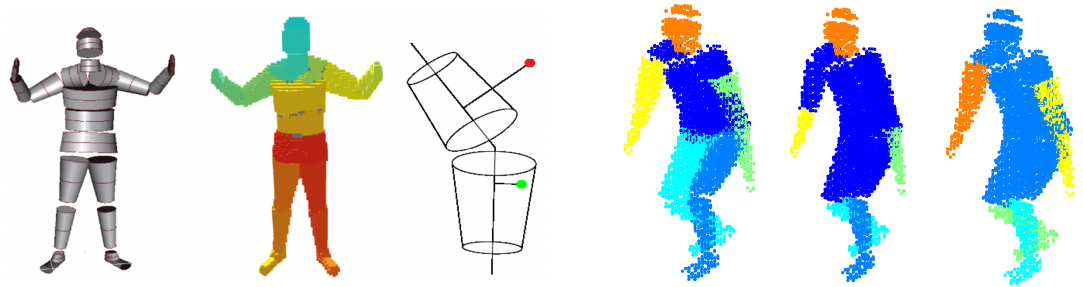}
\caption{Left: Ground-truth labels for voxels are automatically generated from an articulated model of a virtual human. Each voxel is assigned the label of the closest generating link. Right: different performance scores can be calculated by comparing the unsupervised segmentation produced by an algorithm with the three a-priori segmentations shown here (black lines in the plots). \label{fig:ground} }
\end{figure}

In order to quantitatively assess the performance of our segmentation algorithm we produced ground-truth segmentation for the analyzed sequences. In the case of synthetic sequences we used as ground-truth the labels automatically generated by associating each (synthetic) voxel with the closest link of the model which had generated the voxel set in the first place, e.g., Figure \ref{fig:ground}-left.

We worked out three distinct performance measures. Firstly, for a body composed by a number of rigid parts, a segmentation is valid if it does not cut in half a rigid link, i.e., the segmentation is \emph{a coarsening} of the partition of all rigid links. We defined as \emph{coarsening score} the average percentage of points across all rigid links which belong to the majority cluster (the unsupervised cluster containing the largest number of points, whichever it is) for the link.

Secondly, the obtained segmentation can be compared with three different ``natural" subdivisions of the human body, e.g., Figure \ref{fig:ground}-right, by counting for each a priori segment $s$ the histogram distribution $hist_s(l)$ of the different unsupervised labels $l$, and retain the percentage of ``majority" voxels, i.e., those associated with the unsupervised cluster with the highest frequency: $score(s) = \max_l hist_s(l)$. A segmentation score can then be obtained as the simple average score for all segments: $segm = \frac{1}{\# s} \sum_s score(s)$.
\begin{figure}[p!]
\centering
\includegraphics[width=\textwidth]{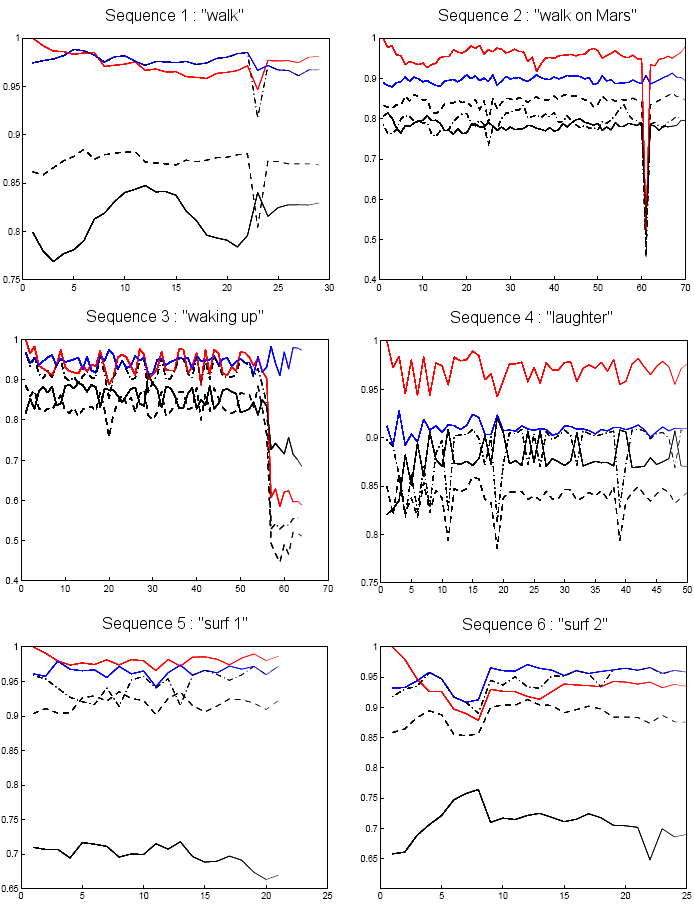}
\caption{Segmentation scores obtained by comparing the labeling generated by our segmentation algorithm with ground truth labels automatically provided for a number of synthetic sequences of articulated motion. Scores obtained over six different sequences of different length from 25 to 70 frames are shown. From top left to bottom right: ``walk", ``walk on mars", ``waking up", ``laughter", ``surf1", and ``surf2". Red: consistency score. Blue: coarsening score. Solid, dash-dot and dashed black lines plot the segmentation scores respectively associated with the three a-priori segmentations of Figure \ref{fig:ground}, taken in the same order. \label{fig:quantitative-synth}}
\end{figure}
When the obtained unsupervised clusters correspond to the a-priori segments the largest cluster contains $100\%$ of the points for each segment, and the score is equal to 1.

Finally, the consistency along time of the obtained unsupervised segmentation is measured by storing at $t=0$ (as a reference) for each link of the body the percentage of voxels which belong to each unsupervised cluster, and measure for each $t>0$ and for each link $l$ the similarity of the current label distribution and the initial one, as 1 minus the $L_1$ distance between the two histograms. This \emph{consistency score} tends to one when the proportion of unsupervised cluster labels is constant in time for all the ground-truth segments of the articulated body.
\begin{figure}[p!]
\centering
\includegraphics[width=\textwidth]{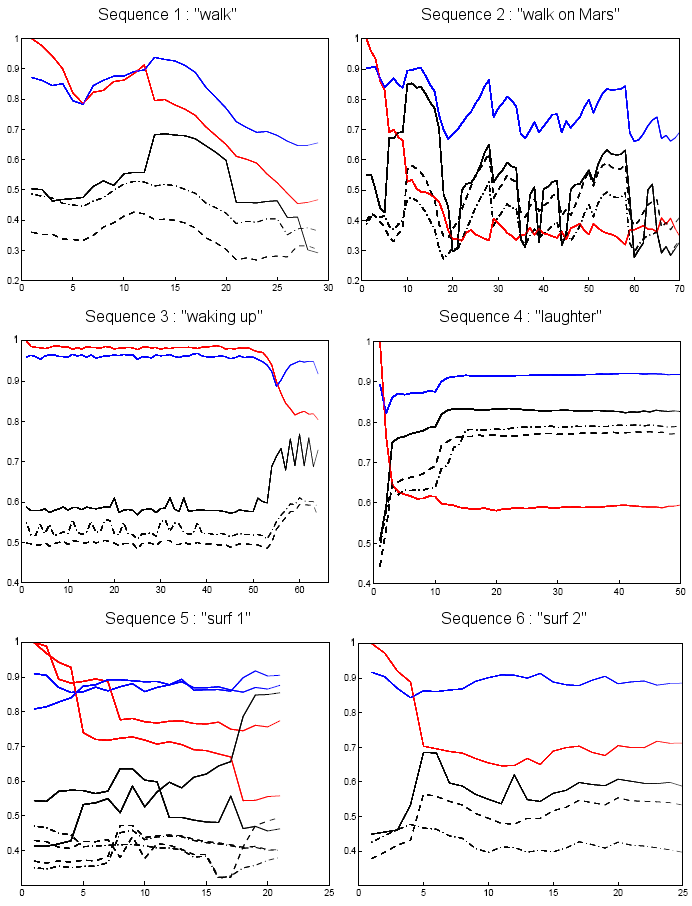}
\caption{Segmentation scores obtained by time-consistent EM clustering on the same synthetic sequences of Figure \ref{fig:quantitative-synth}. For sequence 5, the scores produced by two runs of EM are shown (as there is a random component to it): its performance is consistently not good. \label{fig:em-synth}}
\end{figure}

\subsection{Experiment 1: Quantitative Performance on Synthetic Data} \label{sec:exp1}

Figure \ref{fig:quantitative-synth} shows the segmentation scores for six different synthetic sequences, of length comprised between 25 and 70 frames. We can appreciate how, typically, the obtained unsupervised segmentation turns out to be very consistent along time, as witnessed by a consistency score (in red) between $95\%$ and $100\%$ at all times, even for fairly long sequences picturing quite different types of motion. In all cases the boundaries between unsupervised clusters normally lie in correspondence of actual articulations, as witnessed by the values of the coarsening score (in blue). As for the a-priori segmentation scores (the three different black curves) we can notice that, at least for the given collection of sequences, the designed unsupervised spectral clustering algorithm seems to favor the second a-priori partition of Figure \ref{fig:ground}-right, i.e., it tends to highlight the outermost rigid links rather than whole protrusions (such as legs and arms). This unexpected result can be explained by pointing out that lower density regions (such as articulated joints) cause the embedded cloud to bend.

\begin{figure}[t!]
\centering
\includegraphics[width=\textwidth]{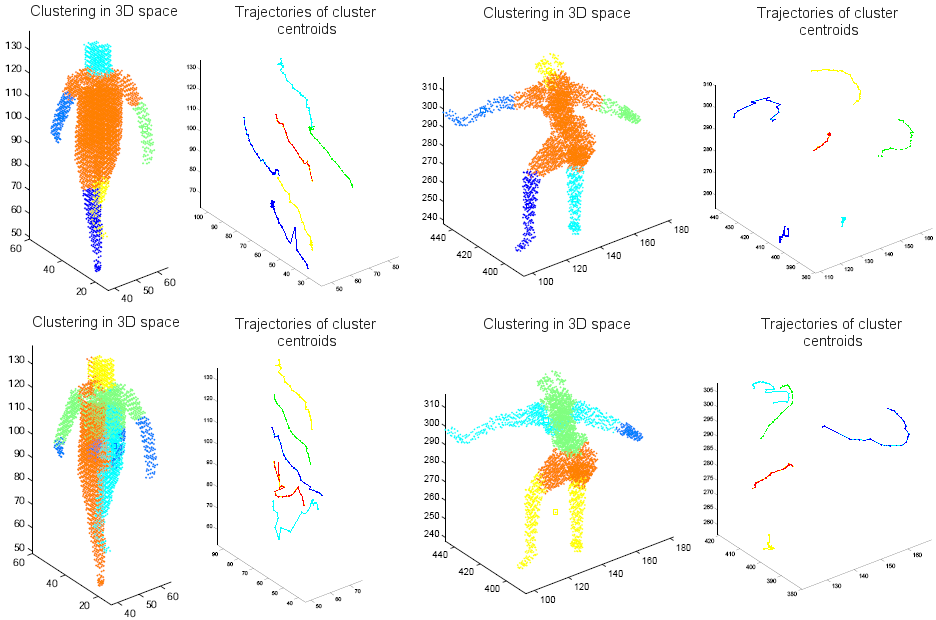}
\caption{Segmentation results (left) and centroid trajectories (right) obtained by our algorithm (top) and time-consistent EM clustering (bottom) for the sequences ``walk" and ``surf1". The voxelset segmentation for the last frame of the sequence is shown as an example. \label{fig:sample-segm}}
\end{figure}

Figure \ref{fig:em-synth} shows the corresponding results produced by time-consistent EM clustering in the original 3D space. Not only the absolute segmentation performance (measured by the black plots) is consistently, dramatically worse than that of the proposed algorithm, but the obtained segmentation is not at all consistent along time (red curves). Incidentally, EM segmentation appears to relatively favor the first a-priori segmentation (whole legs and arms), as the dominance of the solid black curve attests. Clusters drift inside the shape during the motion, and usually span different distinct body-parts. This phenomenon is made clear in Figure \ref{fig:sample-segm}, where the irregular trajectories of the obtained 3D clusters for two sub-sequences of ``walk" and ``surf" are plot, while the obtained segmentation is visually rendered for two key-frames of the two sequences, showing its lack of relation with articulated segments.

\begin{figure}[p!]
\centering
\includegraphics[width=\textwidth]{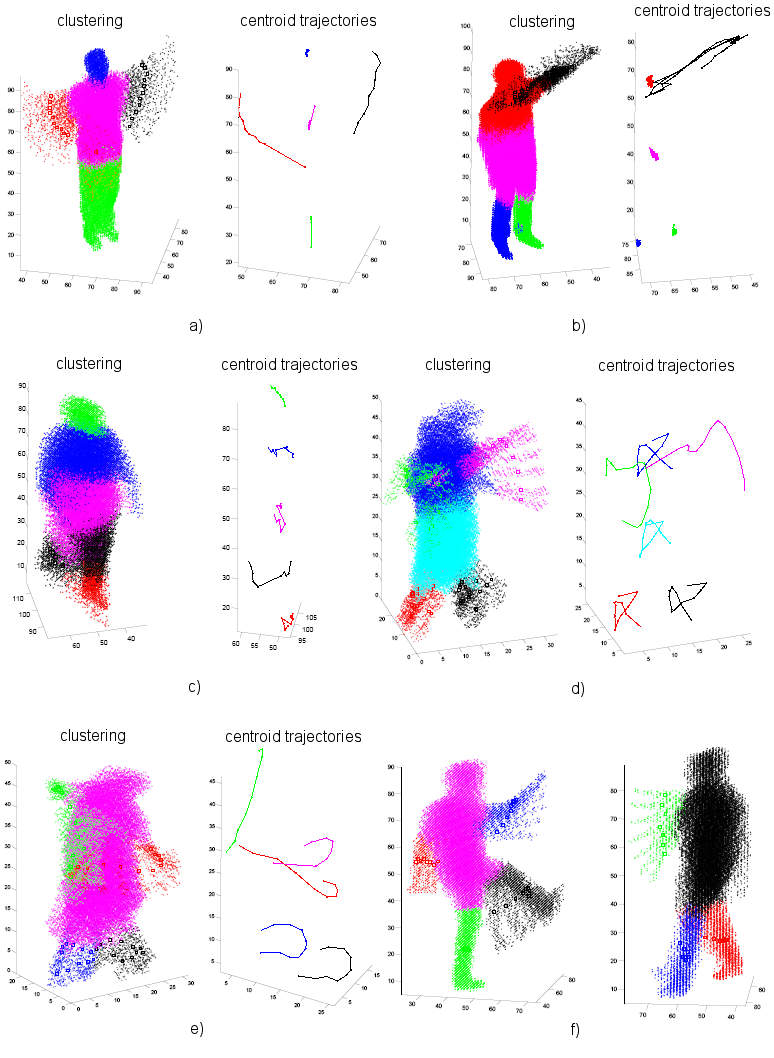}
\caption{Examples of results produced by the dynamic segmentation algorithm of Section \ref{sec:algo} on a number of real-world sequences: a) ``fly" sequence of 11 frames; b) ``arm-waving" sequence of length 50; c) ``walking" motion, length 10; d) a subsequence of ``danceuse", 16 frames; e) another ``danceuse", length 10. Both the evolution of whole unsupervised clusters in 3D and their centroids' trajectories are shown. f) Example of topology transition management: after 11 frames in which arms are spread out
(left) the left arm gets in contact with the torso: the algorithm adapts the number of clusters accordingly and
proceeds to segment in a smooth way for other 7 frames (right, from a different viewpoint). \label{fig:movies}}
\end{figure}

\subsection{Experiment 2: Qualitative Assessment on Real Data} \label{sec:exp2}

In a second series of tests we applied the algorithm of Section \ref{sec:algo} to several different real-world, high-resolution sequences of voxelsets, generated by our multi-camera acquisition system. Among those, a
200-frame-long sequence capturing a dancer who moves and swirls all around the scene, and voxelset resolution of 1cm.

Figure \ref{fig:movies} illustrates some typical results of the dynamic segmentation algorithm described in Section \ref{sec:algo}, for a number of real-world sequences. It can be appreciated how the resulting unsupervised segmentation turns out to be pretty consistent in time, yielding very smooth cluster centroid trajectories, not only in situations where body-parts are well separated ( a), b)) but also during walking gaits (c) or even extremely complicated motions like the dance performed in d), e). Figure \ref{fig:movies}-f) shows how the algorithm copes with transitions in the body's topology, in particular the arm of a walking person touching their torso. Cluster evolution displays remarkable stability before and after the transition.

\subsection{Experiment 3: Quantitative Assessment on Real Sequences} \label{sec:exp3}

When motion capture data are available it is possible to measure quantitatively the performance of the three competing methods (the proposed time-consistent d-wise LLE segmentation, time consistent EM clustering, and time-consistent ISOMAP embedding with K-means clustering) on real-world sequences as well, by assigning as ground-truth to each voxel the label of the closest \emph{estimated} link.

\begin{figure}[t!]
\centering
\includegraphics[width=\textwidth]{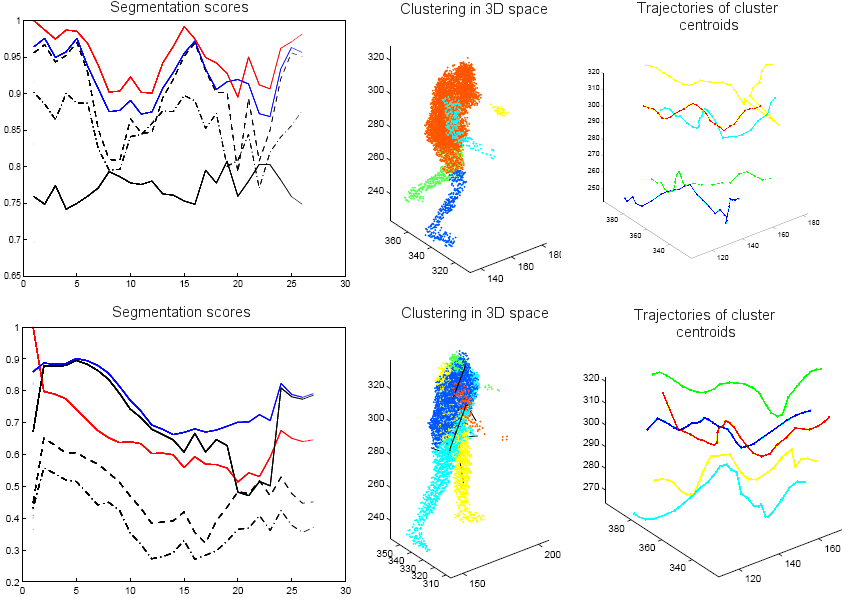}
\caption{Segmentation scores (left) and centroid trajectories (right) obtained by our algorithm for the real-world sequence ``mars", from $t=389$ to $t=415$. Top: our algorithm. Bottom: time-consistent EM clustering. The actual voxelset segmentation corresponding to the critical frame $t=408$ of the sequence is also shown (middle). \label{fig:quantitative-real2}}
\end{figure}

Voxelset sequences actually captured through a multi-camera system suffer from a number of unpleasant phenomena, like presence of large gaps or holes in the grid of voxels (e.g. the gap in the wrists in Figure \ref{fig:quantitative-real2}-middle), noise, disconnected components, not to mention entire missing body-parts (compare the missing head in Figure \ref{fig:quantitative-real2}-middle, top and bottom). Figure \ref{fig:quantitative-real2}-top illustrates how, for adequate values of the parameters (in the plots $d=4, k=25$), the segmentation scores achieved by our algorithm are once again very high, with the method showing remarkable resilience to unreliable data capture (red line). This compares very favorably with the behavior of time-consistent EM clustering (bottom).

We can still notice a couple of drops in the scores (top-left), mirrored by a brief sudden glitch in clusters trajectories, (top-right). They correspond to frames (such as $t=408$, whose voxelset segmentation is shown in the middle) in which gaps are so wide that they affect the quality of the segmentation, even though the shape of the embedding cloud remains stable.
\begin{figure}[t!]
\centering
\includegraphics[width = 1\textwidth]{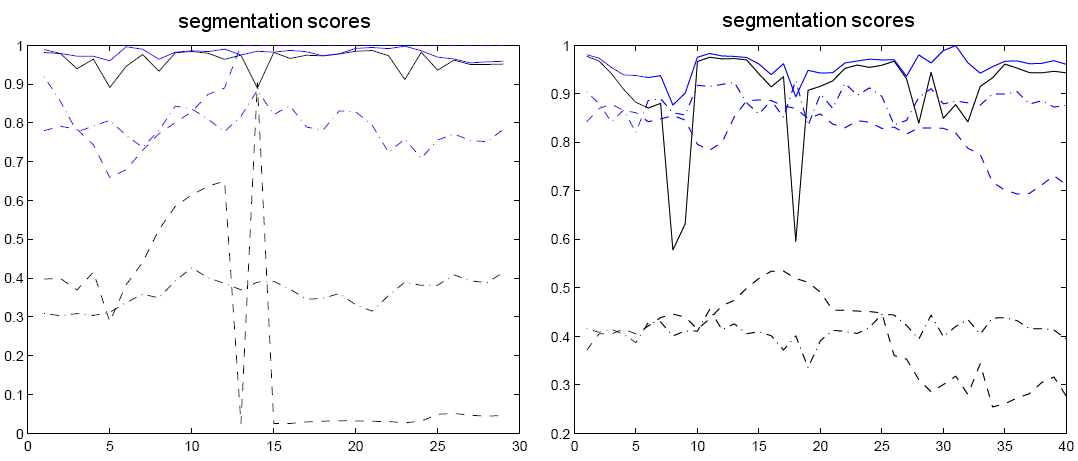}
\caption{Performance scores for two more real-world sequences of length 29 and 40 respectively, and for all the competing methods. Only the coarsening (blue) and the (second) segmentation (in black) scores are plotted. Solid lines: our method, time-consistent d-wise clustering in LLE space; dashed lines: time-consistent EM clustering in 3D. Dash and dotted lines: time-consistent k-means clustering in ISOMAP space. The performance scores of the proposed approach clearly dominate all the others'. \label{fig:quantitative-graphs}}
\end{figure}

Figure \ref{fig:quantitative-graphs} shows the scores obtained by all three competing methods over two other challenging sequences of real voxelsets. It is apparent how our method exhibits strong resilience to data of very poor quality, easily outperforming both EM clustering in 3D or k-means clustering in the geodesic-based ISOMAP space. At times glitches due to extremely corrupted data ($t=8, t=17$, right) appear, but the topology adaptation algorithm of Section \ref{sec:strategy} brings swiftly the segmentation back on track.

\subsection{Sensitivity analysis} \label{sec:exp4}

\subsubsection{Estimating the Optimal Number of Neighbors} \label{sec:k-est}

\begin{figure}[t!]
\centering
\begin{tabular}{c}
\includegraphics[width=\textwidth]{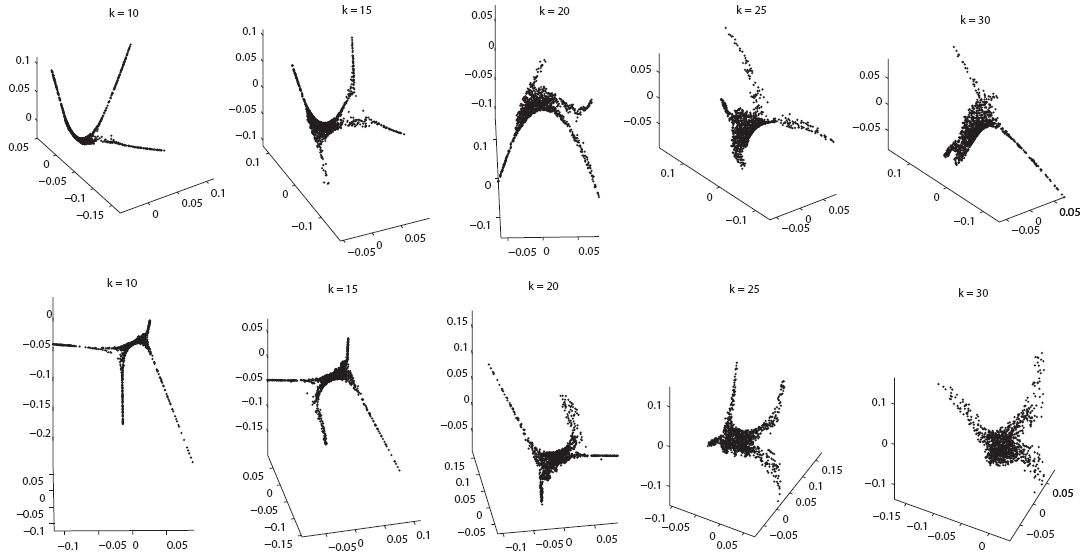} \\ a) \\
\includegraphics[width=\textwidth]{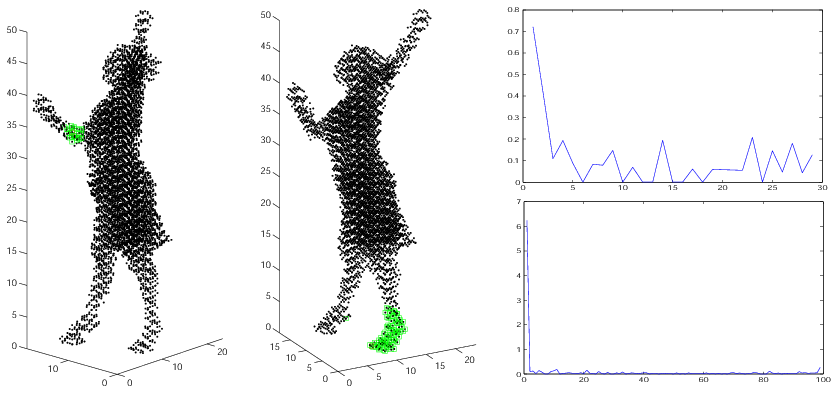} \\ b)
\end{tabular}
\caption{a) This shows how the lower-dimensionality, separability and stability of the embedded cloud along a sequence depend on the size of the neighborhood $k$ in the LLE algorithm. b) Unsuitable values of $k$ are characterized by ``anomalous" neighborhoods which span distinct body-parts (middle), in opposition to the case of admissible values (left). Right: plots of the distance between the farthest point of the neighborhood and all the others in the two cases (top: ``regular" neighborhood; bottom: ``anomalous" neighborhood). \label{fig:sensitivity}}
\end{figure}

The proposed segmentation methodology critically relies on the ``desirable" geometric properties of LLE discussed in Section \ref{sec:lle}, which in turn depend on its two basic parameters: the size $k$ of the neighborhoods and the dimension $d$ of the embedding space. The former, in particular, affects both the stability of the embedded shape along time and its lower-dimensionality, from which the estimation of the number of clusters itself depends. It can be noticed empirically that, while the embedded shape shows a remarkable stability for some values of $k$, this is not in general the case for arbitrary such values. Figure \ref{fig:sensitivity}-a) shows how the embedded shape varies for two frames (top, bottom) in the ``dancer" sequence as $k$ ranges from 10 to 30.

For higher values of $k$ some neighborhoods of points in a given body-part comprise regions of a different body-part, e.g., Figure~\ref{fig:sensitivity}-b). However, in those ``anomalous" neighborhoods the farthest element (as it belongs to another, distinct link) is relatively distant from all others. If we plot the distance between this point and all its fellows we can notice a large jump, e.g.,Figure \ref{fig:sensitivity}-b) (bottom right). This is not the case for ``regular" neighborhoods spanning a single rigid part (top right). We can then set as acceptable value for $k$ any of those  which yield only ``regular" neighborhoods.

\subsubsection{Laplacian Methods, Selection of Eigenfunctions and Body-part Resolution} \label{sec:selection}

Spectral methods have long been used in the computer graphics community as a tool to analyze and process meshes representing surfaces of objects inside virtual scenes. Generally speaking, they share a common framework in which an affinity matrix $M$ which reflects the structure of the input set of points is defined and an eigen decomposition of this matrix which yields its eigenvalues and eigenvectors is performed. They can be roughly classified into two categories: methods in which $M$ encodes adjacency information (such as LLE or ISOMAP), and methods based on the \emph{graph Laplacian}, an operator $\mathcal{L} : f \mapsto \mathcal{L} f$ mapping functions $f : X \rightarrow \mathbb{R}$, $X_i \mapsto f(X_i) = f_i$ defined on sets of points (vertices) $X = \{X_i, i = 1,...,N\}$ forming a graph, of the form:
\begin{equation} \label{eq:laplacian}
(\mathcal{L} f)_i \propto \sum_{j \in N(i)} w_{ij} \cdot (f_i - f_j)
\end{equation}
where $N(i)$ is the set of neighbors of the point $X_i$ (the vertices connected to it by an edge) in $X$, and $w_{ij}$ is the weight of the edge joining $X_i$ with $j$-th neighbor.
\begin{figure}[t!]
\centering
\includegraphics[width = \textwidth]{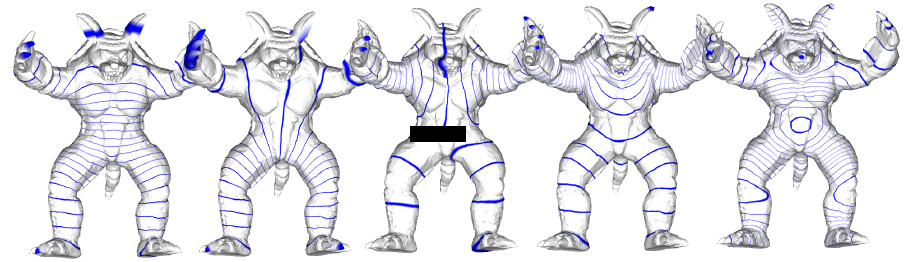}
\caption{As stationary functions on the shape approximated by a cloud of points, Laplacian eigenfunctions are associated with symmetries and protrusions of the underlying domain. Here the level sets (in blue) of the first five eigenfunctions of the depicted ``monster" are shown to partition the toy along directions that follow its natural axes of symmetry: horizontal (leftmost image), left/right (second picture), or radiating from the barycenter (last image). \label{fig:understand}}.
\end{figure}

Functions on a set $X$ of $N$ points can be trivially represented as the vector of their values on the $N$ points: $f = [f(X_i) = f_i]^T$. Therefore, the $N$ by $N$ affinity matrix $M$ computed by LLE can also be seen as an operator on the space of functions defined on the cloud of $N$ points $X$, as the matrix multiplication $f' = M \times f$ yields another vector $f'$ of $N$ components, representing another function on $X$. As it can be proven that the affinity matrix operator can be approximated by the square Laplacian, $M f = (I-W)^T (I-W) f \approx \frac{1}{2} \mathcal{L}^2 f$, Locally Linear Embedding can be considered a form of Laplacian embedding.

Now, the level sets of Laplacian eigenfunctions are related to the geometry of the underlying shape $X$. In particular \emph{nodal sets} \cite{toth01geometric}, i.e., zero-level sets of eigenfunctions, partition $X$ into a set of ``nodal domains" strongly related to protrusions and symmetries of the underlying grid of points.\\
Figure \ref{fig:understand} shows a clear example of how the level sets of different eigenfunctions of the Laplacian operator defined on (the discretized version of) any given shape follow the symmetries of the shape itself \cite{levi06smi}. As the $n$-th eigenfunction has at most $n$ nodal domains, the first eigenfunctions are associated with relatively ``coarse" partitions of the shape of interest.
\begin{figure}[t!]
\centering
\includegraphics[width=\textwidth]{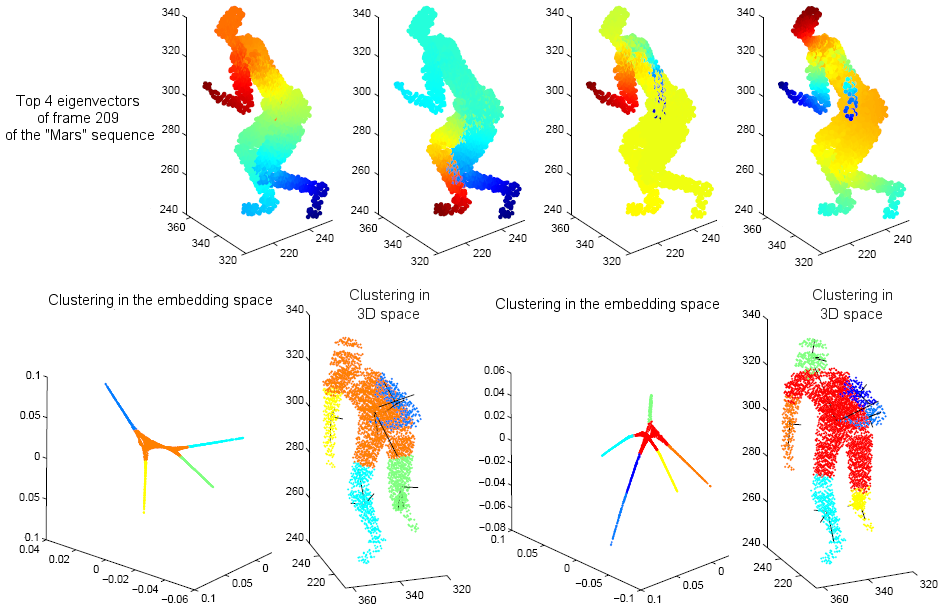}
\caption{Different eigenfunctions capture different aspects of the shape's geometry. Top: the first four eigenfunctions for the frame $t=209$ of the ``mars" sequence. Each eigenfunction corresponds to a 1 by $N$ eigenvector of the affinity matrix: its value of the $i$-th point of the grid is the $i$-th entry of the eigenvector and is rendered here in color. Bottom-left: segmented embedding cloud and corresponding 3D segmentation for the same frame, obtained by selecting eigenvectors 1,2, and 3. Bottom-right: segmentation associated with eigenvectors 2,3, and 4. \label{fig:eigenfunctions}}
\end{figure}

As the LLE affinity matrix is linked to the graph Laplacian operator, its eigenvectors are also associated with specific symmetries of the underlying cloud of points. In other words, the choice of which eigenvectors of $M$ we select after SVD determines the structure of the embedded cloud. While in standard LLE always the bottom $d$ are chosen, $d$ being the dimension of the resulting embedding, LLE is used here just as a tool to make the structure of the 3D shape in terms of protrusions clearly emerge. It makes then more sense to look for an appropriate, arbitrary \emph{selection of eigenfunctions} of the affinity matrix able to facilitate the clustering of the shape at hand.

To visually render these concepts, we can visualize functions $f$ (and in particular eigenvectors of $M$) on the original cloud $X$ of 3D points as ``colored" versions of the same cloud, in which the color of each point $X_i$ is determined by the value $f(X_i)$ of the function in that particular point.

We used this technique to depict in Figure \ref{fig:eigenfunctions}-top the top four eigenvectors of the LLE affinity matrix for frame 209 of the ``Mars" sequence. It is striking (but to be expected, following the brief discussion on graph Laplacian methods) to notice how the positive (red) and negative (dark blue) peaks of the different eigenfunctions of $M$ are located exactly on the protrusions of the underlying shape, i.e., its high-curvature regions. Selecting a specific set of eigenfunctions is then equivalent to highlighting the associated peaks.\\ The resulting embedded clouds may even possess a different number of distinguishable branches. In Figure \ref{fig:eigenfunctions}-bottom-left, for instance, eigenfunctions 1,2,3 determine an embedding not able to resolve the head as a separate protrusion. The head appears instead as an additional cluster when selecting eigenfunctions 2,3,4 (right).
\begin{figure}[t!]
\centering
\includegraphics[width=\textwidth]{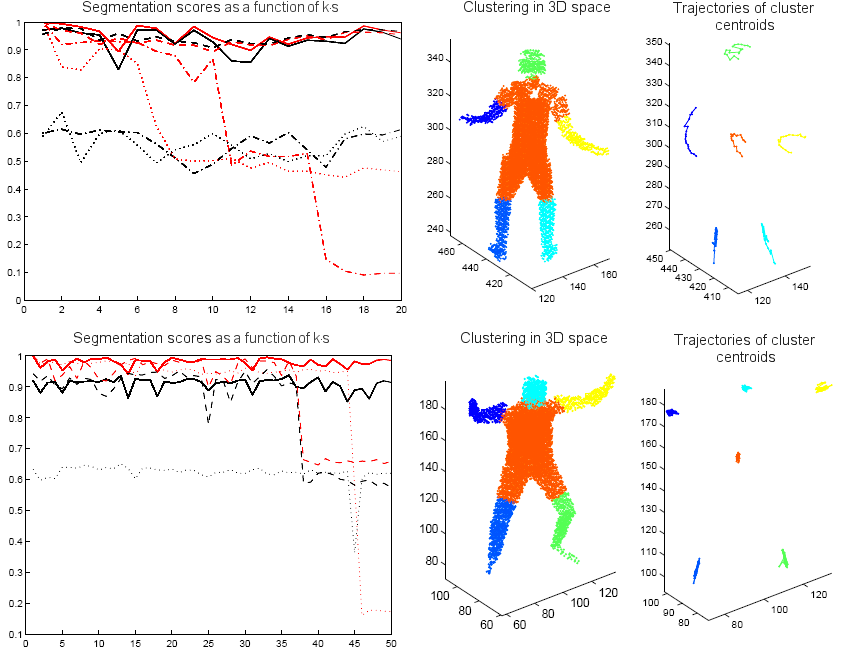}
\caption{How the discrete sampling of a 3D shape influences the performance of the algorithm. Top: sequence ``space surf" from $t=300$ to $t=319$. Bottom: ``reveil" from $t=150$ to $t=200$. Left: consistency (red) and segmentation (black) scores throughout each sequence, for different values of the product $k$ by $s$: $3 \times 14$ (solid), $4 \times 10$ (dashed), $5 \times 8$ (dash-dotted), $6 \times 7$ (dotted) for ``space surf"; $5 \times 18$, $6 \times 15$ and $7 \times 14$ for ``reveil". Center: example segmentation  ($t=319$ for $s=3$, $k = 14$ for the top sequence; $t=218$ for the bottom one). Right: corresponding centroid trajectories. \label{fig:sampling}}
\end{figure}

\subsubsection{Influence of Voxel-grid Resolution} \label{sec:sampling}

As we just argued, the quality of the segmentation depends on the ``good" features of the embedding, namely its lower dimensionality and improved branch separation. In turn, those depend (amongst other factors) on the assumption that \emph{the graph Laplacian of the actual cloud of points representing the shape is a good approximation of the (unknown) Laplace-Beltrami operator}.

Figure \ref{fig:sampling} illustrates how performances degrade as the number of voxels (seen as samples of an underlying continuous shape) decreases. To allow a fair comparison we kept roughly constant at all runs the product $s \cdot k$ between number of neighbors $k$ and sampling factor $s$ (as the original voxel sets contain as many as 25000 voxels, it is convenient to sub-sample them to limit computation time). In Figure \ref{fig:sampling} the consistency score and the segmentation score for the ``natural" segmentation of the body into lower limbs, head, and bulk are shown for two subsequences of ``space surf" and ``reveil" (top and bottom, respectively). For more static sequences (bottom) the segmentation score (black) is more stable as a function of the product $k \cdot s$, even though subject to sudden jumps. The consistency score (red) is more sensitive, while performance degrades consistently when grid resolution is reduced for sequences in which the body moves significantly (top).

\begin{figure}[t!]
\centering
\includegraphics[width=\textwidth]{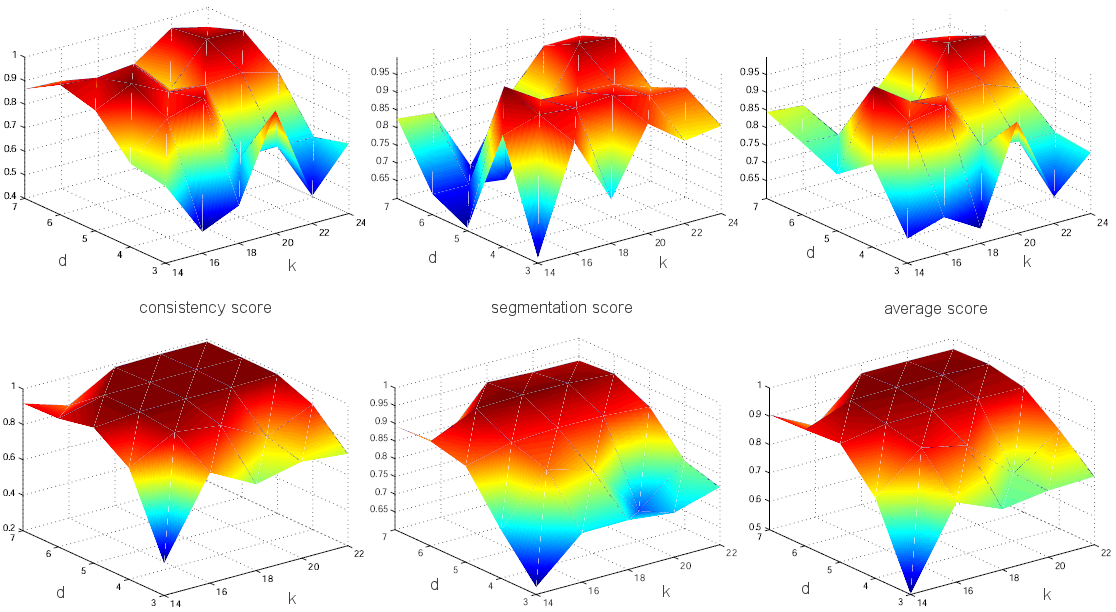}
\caption{Consistency (left), segmentation (middle), and average (right) scores obtained over two sample synthetic sequence (top and bottom) for different values of the parameters $k = \{14,16,18,20,22\}$ (on the abscissa) and $d = \{3,4,5,6,7\}$ (on the ordinate) of the algorithm. Top performances are achieved for a wide range of the parameters. Top: ``walk", from $t=71$ to $t=80$. Bottom: ``surf", from $t=1$ to $t=25$. \label{fig:selection}} 
\end{figure}

This is due to two different effects. On one side, for sparser clouds of points the graph Laplacian is a worse approximation of the Laplace-Beltrami operator. On the other, as the list of points is randomly sampled to produce the reduced dataset, the fewer the samples the less they happen to be distributed along the nodes of a regular 3D grid, as in the original voxel set. As this deforms the local structure of the neighborhoods, it causes irregular instability in the embedded cloud along the sequence which influences in particular the consistency score (red).
\begin{figure}[t!]
\centering
\includegraphics[width=0.8\textwidth]{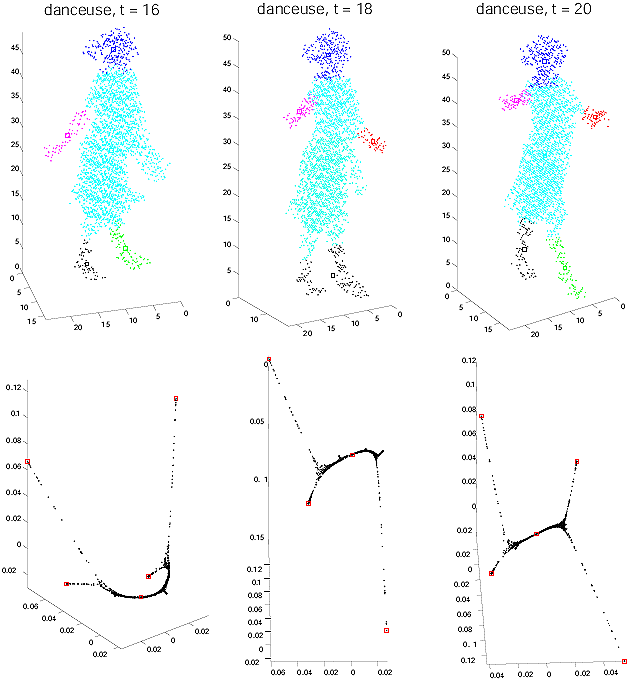}
\caption{How the splitting/merging algorithm deals with topology changes in the embedded space. At $t=16$ the
left arm of the dancer touches her scarf (left), and a single cluster covers body, arm, and scarf. Then ($t=18$) the
left arm becomes visible and a new cluster is assigned to it, while the dancer's feet get too close to each other to be distinguished (middle). Finally ($t=20$) her legs pull again apart, generating a separate cluster for each of them (right).
\label{fig:change}}
\end{figure}

\subsubsection{Robustness with Respect to the LLE Parameters}

It is important to assess the sensitivity of the algorithm to the main parameters $k$ and $d$ (or, better, the indices of the selected eigenvectors). Figure \ref{fig:selection} illustrates how the segmentation scores vary when different values of the parameters $k,d$ are used to compute the embedding (assuming for sake of simplicity that we want to select the first $d$ eigenfunctions). The consistency score, the segmentation score (w.r.t. the second a-priori segmentation of Figure \ref{fig:ground}-right) and their average are plotted as functions of $k$ and $d$, for $k = \{14,16,18,20,22\}$ and $d = \{3,4,5,6,7\}$, for two synthetic subsequences of ``walk" and ``surf". The stability of both consistency in time and quality of the segmentation in a fairly large region of the parameter space is apparent, and speaks for the robustness of the approach.

\subsection{Robustness to Topological Changes} \label{sec:exp5}

No matter how robust to pose variation LLE embedding may be, instants in which different parts of the articulated body come into contact still have important effects on the shape of the embedded cloud. These events have truly dramatic consequences on embeddings based on measuring geodesic distances along the body, since new paths appear, affecting in general the distance between \emph{all} pairs of points in the cloud.
\begin{figure}[!]
\centering
\includegraphics[width=\textwidth]{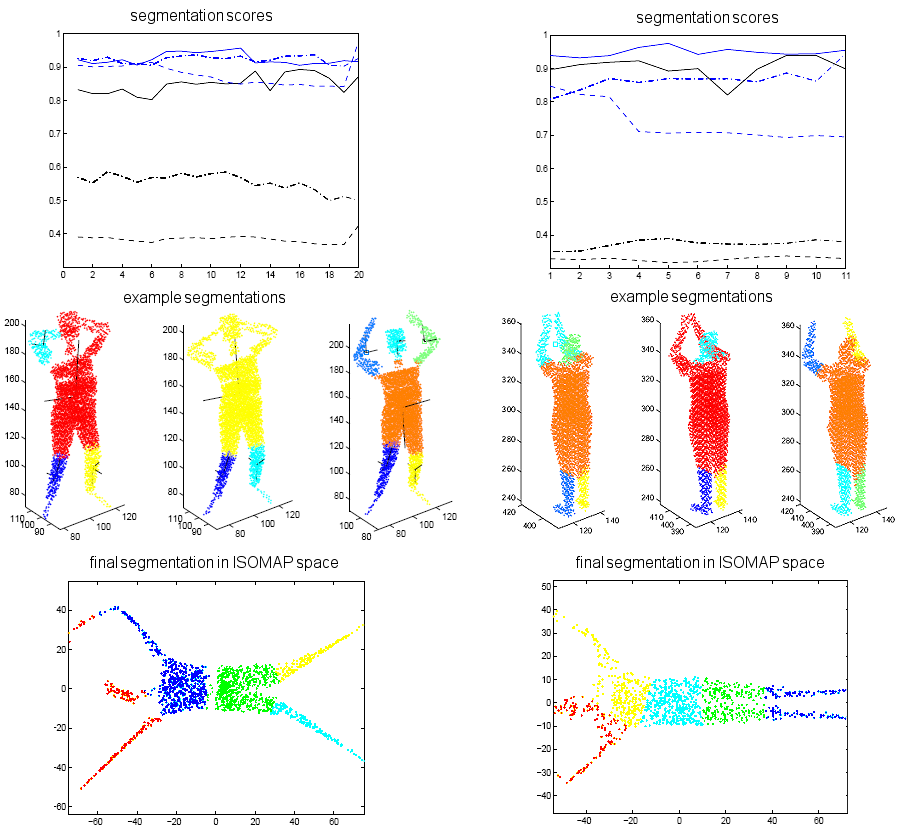}
\caption{Measuring the relative performance of ``local" methods based on graph Laplacians (represented in this paper by LLE), EM clustering, and ``global" embeddings based on geodesic distances (represented by ISOMAP) for sequences affected by topology changes. Left column: ``wake up" sequence, from $t=95$ to $t=114$. Right column: ``clap" sequence. Top: consistency (blue) and segmentation (black) scores throughout the sequence: solid - our method; dashed - time-consistent EM clustering; dash-dotted - time-consistent k-means clustering in ISOMAP space. Middle: some examples of how our algorithm copes with topology transitions are given in terms of 3D segmentations before, during and after the transition. Bottom: ISOMAP clusterings for the final frames of the two sequences. \label{fig:change-comp}}
\end{figure}

Figure \ref{fig:change} illustrates how the splitting-merging algorithm of Section \ref{sec:strategy} copes with such changes. Full consistency of the segmentation cannot be preserved, as the number of protrusions in the embedding shape changes, and we do not make use of point-to-point correspondences (which would themselves have trouble coping with such situations). Nevertheless, most protrusion end up been preserved through the event changing the body's topology, while the others can be recovered as soon as they can be distinguished once again.

Figure \ref{fig:change-comp}, instead, compares the segmentation scores of methods based on the local (LLE) and global (ISOMAP) structure of the shape in similar situations, in which the topology of the body changes, for two significant synthetic sequences (``wake up", left, and ``clap", right). As the plots on top show, propagating clusters in the LLE space exhibits superior performance and robustness, as the algorithm smoothly adapts to topology changes in virtue of the geometric features of LLE embedding.

\begin{figure}[t!]
\centering
\includegraphics[width=\textwidth]{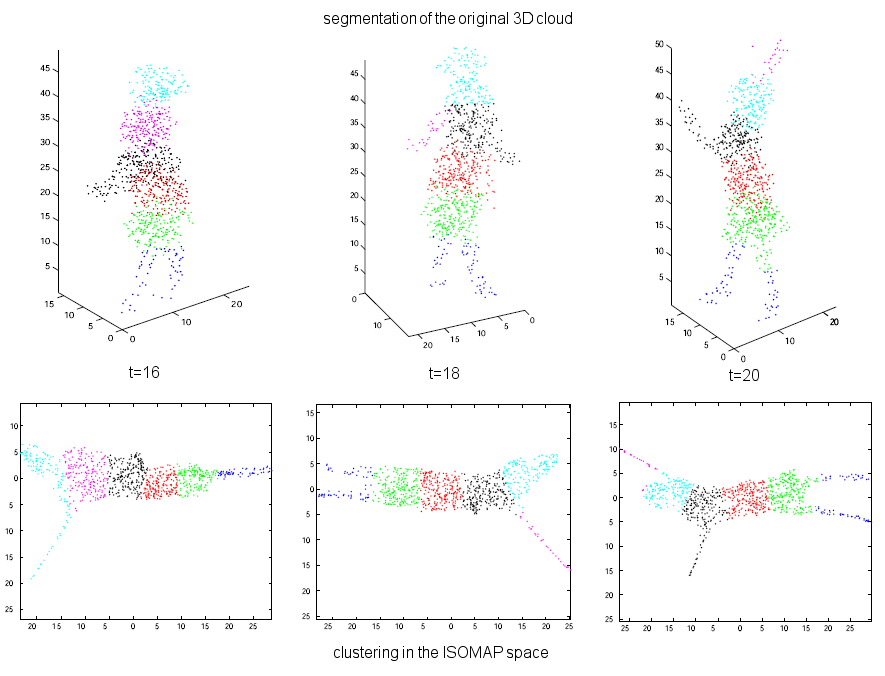}
\caption{Behavior of ISOMAP along the sequence of topology transitions of Figure \ref{fig:change}. The shape of the
embedded cloud (we chose $d=2$ for sake of readability) changes dramatically, gravely affecting the segmentation in the original 3D space. \label{fig:change-iso}}
\end{figure}

The bottom plots in Figure \ref{fig:change-comp} show, as an example, the segmentations produced by k-means in the ISOMAP space for the final frame of the two sequences. As ISOMAP embeddings do not exhibit any detectable lower-dimensionality or separation between body-parts / protrusions, k-wise clustering is just not feasible there. Clustering is then performed in the ISOMAP space by k-means. Indeed, Figure \ref{fig:change-iso} illustrates how ISOMAP (as a representative of all geodesic-based spectral methods) copes with the same transitions of Figure \ref{fig:change}. The comparison clearly shows the advantage of doing clustering after Laplacian embedding.

\section{An Application to Bottom-up Model/Pose Recovery} \label{sec:developments}

\begin{figure}[t!]
\centering
\includegraphics[width=\textwidth]{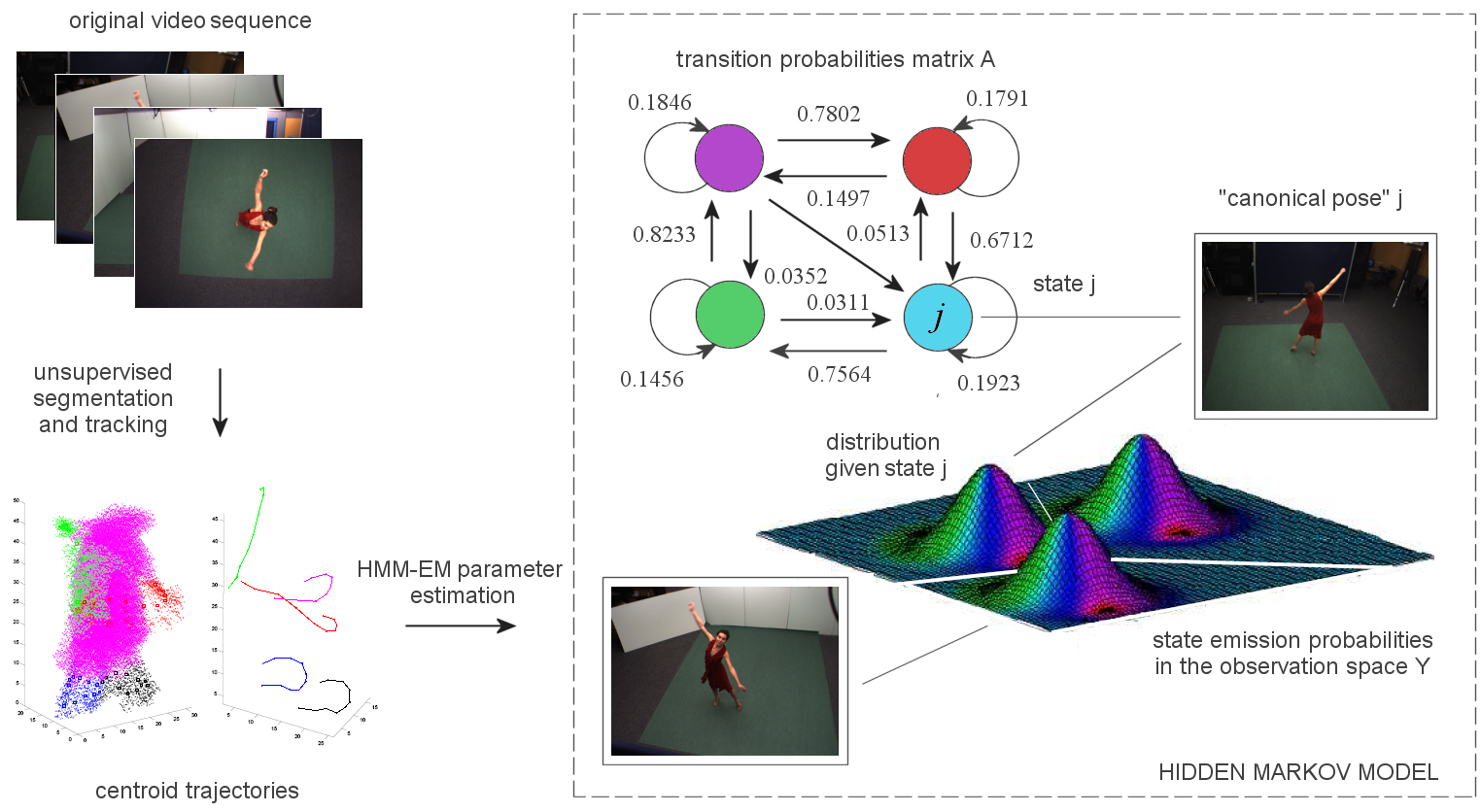}
\caption{The segments' centroid trajectories produced by our unsupervised segmentation algorithm are consistent enough to be used as feature vectors for a parameter estimation algorithm generating a dynamical model (for instance, a hidden Markov model) which represent the input video sequence. Such a model can later be used for action classification. \label{fig:action-recognition}}
\end{figure}

In our view, the unsupervised segmentation algorithm we propose can be seen as a building block of a wider motion analysis framework. We mentioned in the introduction two possible applications of time-consistent unsupervised segmentation: action recognition and bottom-up body-model recovery. As for the first one, a look at Figures \ref{fig:sample-segm}, \ref{fig:movies}, and \ref{fig:quantitative-real2} clearly shows the smoothness and coherence of the tracks associated with the centroids of the unsupervised clusters generated by our framework. Just as in the case of image features, we can assume that these tracks are generated by a graphical model, such as for instance an HMM \cite{Moore95} (see Figure \ref{fig:action-recognition}), or possibly a sophisticated hierarchical, non-linear \cite{ralaivola04dynamical}, or even chaotic \cite{ali07chaotic} model. Classical algorithms can then be employed to estimate the parameters of this model in a generative approach to recognition: each test sequence is then classified according to the label of the training model which is the closest to the one that more likely generated it.

Another natural application of our unsupervised, time-consistent segmentation algorithm is recovering and fitting simple stick models to the segmented clusters along the sequence. As it provides a coherent protrusion segmentation along a sequence, and protrusions typically correspond to chains of rigid links, its output is suitable to reconstruct rough models of the moving body as a first step, for instance, of a model-free motion capture algorithm. However, here we do not aspire at providing a full solution to this problem. We can illustrate with a simple example of how this can be done. Ellipsoids can be easily fitted to the segmented protrusions, for instance by aligning the moments or principal axes, or position sticks along the main axis of the segmented body-part.

\begin{figure}[t!]
\centering
\includegraphics[width=\textwidth]{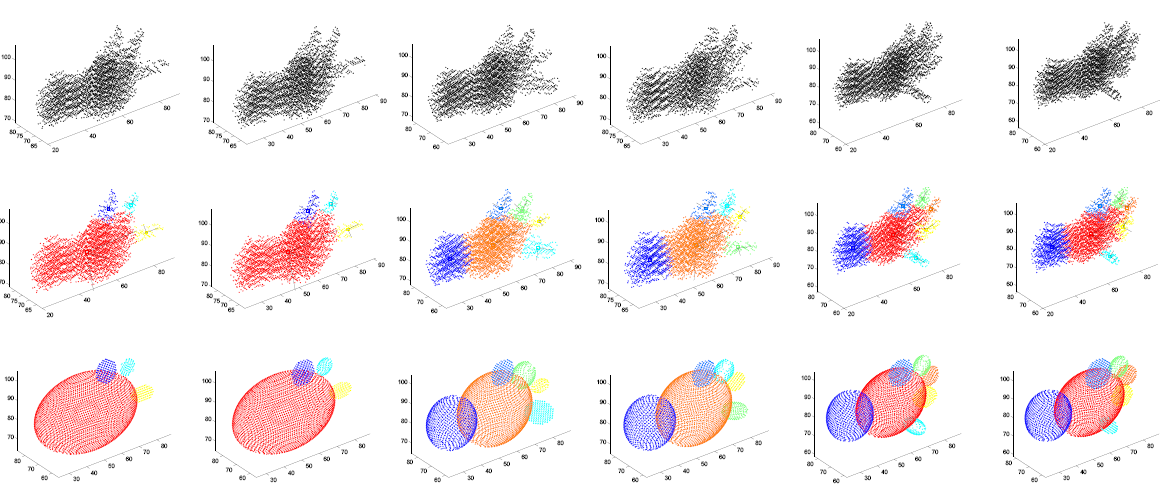}
\caption{Fitting a rough model formed by a number of ellipsoids to the obtained clusters is straightforward and yields remarkable results. Top: a sequence of voxel-sets capturing a hand's motion in an augmented reality environment where the user interacts with virtual objects. Bottom: corresponding rough articulated model fitting. \label{fig:hand}}
\end{figure}

Figure \ref{fig:hand} shows the resulting model fitted to a sequence of voxel-sets representing a hand and its fingers. In an augmented reality environment the 3D reconstruction of a hand could be used to interact with virtual objects in a virtual environment. \addnote[virtual]{1}{From this ellipsoid-based representation one can easily infer an implicit-surface representation which can be used for real-time interactive applications.  Such a parametric description of the object is complementary to data, e.g., voxel, representations.}
It can be noted by looking at Figure \ref{fig:hand} that our methodology does not (by construction) necessarily identify all articulations. It does, however, identify remarkable protrusions in the moving body. As the hand evolves and more fingers become clearly separated from the palm, they are isolated as separated clusters and the rough model of the hand can be updated. Even though limb articulations do not show up in the embedding space, these can be later identified by refining the obtained segments in a more comprehensive model recovery framework. 

\section{Conclusions} \label{sec:conclusions}

In this paper we presented a novel dynamic, unsupervised spectral segmentation scheme in which moving articulated bodies are clustered in an embedding space, and clusters are propagated in time to ensure temporal consistency. By exploiting some desirable geometric characteristics of LLE, in particular, we can estimate the optimal number of clusters in order to merge/split them when topology transitions occur. We compared the performance of our algorithm versus (time-consistent) EM clustering in 3D, k-means clustering in ISOMAP space, and ground truth labeling provided through motion capture or synthetic generation.

An extension of the proposed approach to widely separated, non contiguous poses can be devised in a quite straightforward manner by resorting for cluster propagation to methods that match different poses of the same articulated object by aligning their embedded images \cite{jain06smi,Bronstein06generalized,Mateus08cvpr}. Though the methodology has been proposed for articulated objects, we can also hope to extend it to certain classes of deformable objects. This could be done by exploiting the property of graph Laplacian methods which states that the eigenvalues of the graph Laplacian remain stable under deformations of the object to segment which preserve its volume \cite{vgl}. Much further work needs to be done in this sense, and will be the aim of our research in the near future.



\begin{thebibliography}{}

\bibitem{bpc}
S.~Agarwal, J.~Lim, L.~Zelnik-Manor, P.~Perona, D.~Kriegman and S.~Belongie.
\newblock Beyond pairwise clustering.
\newblock In {\em Computer Vision and Pattern Recognition}, 838 -- 845, 2005.

\bibitem{ali07chaotic}
S.~Ali, A.~Basharat, and M.~Shah.
\newblock Chaotic invariants for human action recognition.
\newblock In {\em International Conference on Computer Vision}, 2007.

\bibitem{Alpert99spectral}
A.~B. Alpert, C.~J.~Kahng and S.-Z. Yao.
\newblock Spectral partitioning with multiple eigenvectors.
\newblock In {\em Discrete Applied Mathematics}, 90(1--3):3--26, 1999.

\bibitem{Bronstein06generalized}
A.~M. Bronstein, M.~M. Bronstein and R.~Kimmel.
\newblock Generalized multidimensional scaling: A framework for
  isometry-invariant partial surface matching.
\newblock In {\em Proceedings of the National Academy of Sciences},
  103(5):1168--1172, 2006.

\bibitem{Bronstein09ijcv}
A.~M. Bronstein, M.~M. Bronstein and R.~Kimmel.
\newblock Topology-invariant similarity of nonrigid shapes.
\newblock In {\em International Journal of Computer Vision}, 81(3):281--301,
  2009.

\bibitem{Belkin03laplacian}
M.~Belkin and P.~Niyogi.
\newblock Laplacian eigenmaps for dimensionality reduction and data
  representation.
\newblock In {\em Neural Computation}, 15:1373--1396, 2003.

\bibitem{Bengio04learning}
Y. Bengio, O. Delalleau, N. Le Roux, J.-F. Paiement, P. Vincent and M. Ouimet.
\newblock Learning eigenfunctions links spectral embedding and kernel {PCA}.
\newblock In {\em Neural Computation}, 16(10):2197--2219, 2004.

\bibitem{bengio03}
Y.~Bengio, J.-F. Paiement and P.~Vincent.
\newblock Out-of-sample extensions for {LLE}, {I}somap, {MDS}, eigenmaps, and
  spectral clustering.
\newblock Technical report, Universit\'e Montreal, 2003.

\bibitem{bissacco07pami}
A.~Bissacco, A.~Chiuso, and S.~Soatto.
\newblock Classification and recognition of dynamical models: The role of
  phase, independent components, kernels and optimal transport.
\newblock In {\em IEEE Transactions on Pattern Analysis and Machine Intelligence}, 29(11):1958--1972, 2007.

\bibitem{Biyikoglu07book}
J.~Biyikoglu, T.~Leydold and P.~F. Stadler.
\newblock {\em Laplacian Eigenvectors of Graphs}.
\newblock Springer, 2007.

\bibitem{action-irani-spacetime-shape-iccv05}
M.~Blank, L.~Gorelick, E.~Shechtman, M.~Irani, and R.~Basri.
\newblock Actions as space-time shapes.
\newblock In {\em International Conference on Computer Vision},1395--1402, 2005.

\bibitem{brand99a}
M.~Brand.
\newblock Shadow puppetry.
\newblock In {\em International Conference on Computer Vision}, 1999.

\bibitem{brostow04a}
G.-J. Brostow, I.~Essa, D.~Steedly and V.~Kwatra.
\newblock {Novel skeletal representation for articulated creatures}.
\newblock In {\em European Conference on Computer Vision}, 2004.

\bibitem{bb69593}
N.~Carter, D.~Young, and J.~Ferryman.
\newblock Supplementing {M}arkov chains with additional features for
  behavioural analysis.
\newblock In {\em Proceedings of AVSBS}, 2006.

\bibitem{Chang08automatic}
W.~Chang and M.~Zwicker.
\newblock Automatic registration for articulated shapes.
\newblock In {\em Eurographics Symposium on Geometry
  Processing}, 2008.

\bibitem{chaudry09histograms}
R.~Chaudhry, A.~Ravichandran, G.~Hager and R.~Vidal.
\newblock Histograms of oriented optical flow and {B}inet-{C}auchy kernels on
  nonlinear dynamical systems for the recognition of human actions.
\newblock In {\em Computer Vision and Pattern Recognition}, pages 1932--1939, 2009.

\bibitem{jenkins03cvpr}
C.-W. Chu, O.~C. Jenkins and M.~J. Mataric.
\newblock Markerless kinematic model and motion capture from volume sequences.
\newblock In {\em Computer Vision and Pattern Recognition}, pages 475--482, 2003.

\bibitem{Chung97book}
F.~Chung.
\newblock {\em Spectral Graph Theory}.
\newblock American Mathematical Society, 1997.

\bibitem{cuzzolin04icip}
F.~Cuzzolin, A.~Sarti, and S.~Tubaro.
\newblock Action modeling with volumetric data.
\newblock In {\em Proceedings of ICIP}, volume~2, pages 881--884, 2004.

\bibitem{cuzzolin04siena}
F.~Cuzzolin, A.~Sarti, and S.~Tubaro.
\newblock Invariant action classification with volumetric data.
\newblock In {\em International Workshop on Multimedia Signal Processing},
  pages 395--398, 2004.


\bibitem{cuzzolin07workshop}
F.~Cuzzolin, D.~Mateus, E.~Boyer, and R.~Horaud.
\newblock Robust spectral 3D-bodypart segmentation along time
\newblock in {Workshop on Human Motion -- Understanding, Modeling, Capture and Animation},
196--211, 2007.

\bibitem{cuzzolin08cvpr}
F.~Cuzzolin, D.~Mateus, D.~Knossow, E.~Boyer, and R.~Horaud.
\newblock Coherent Laplacian 3-D protrusion segmentation
\newblock In {Computer Vision and Pattern Recognition}, 2008.

\bibitem{Cvetkovic98book}
M.~Cvetkovic, D. M.~Doob and H.~Sachs.
\newblock {\em Spectra of Graphs: Theory and Applications}.
\newblock Vch Verlagsgesellschaft Mbh, 1998.

\bibitem{Anguelov04nips}
D. Anguelov, P. Srinivasan, D. Koller, S. Thrun, H. Pang and J. Davis.
\newblock The correlated correspondence algorithm for unsupervised registration
  of non-rigid surfaces.
\newblock In {\em Neural Information Processing Systems}, 2004.

\bibitem{Anguelov04recovering}
D. Anguelov, D. Koller, H.-C. Pang, P. Srinivasan and S. Thrun.
\newblock Recovering articulated object models from 3D range data.
\newblock In {\em Proceedings of UAI}, 2004.

\bibitem{dempster77em}
A.~P. Dempster, N.~M. Laird and D.~B. Rubin.
\newblock Maximum likelihood from incomplete data via the em algorithm.
\newblock In {\em J. Royal Stat. Soc.}, 39:1--38, 1977.

\bibitem{Dhillon04kernel}
Y.~Dhillon, I. S.~Guan and B.~Kulis.
\newblock Kernel kmeans: spectral clustering and normalized cuts.
\newblock In {\em Proceedings of KDD}, pages 551--556, 2004.

\bibitem{elgammal04}
A.~Elgammal and C.~Lee.
\newblock {Inferring 3D body pose from silhouettes using activity manifold
  learning}.
\newblock In {\em Computer Vision and Pattern Recognition}, 2004.

\bibitem{Moore95}
R.~Elliot, L.~Aggoun, and J.~Moore.
\newblock {\em Hidden Markov models: estimation and control}.
\newblock Springer Verlag, 1995.

\bibitem{Fouss07random}
A.~R.-J.-M. Fouss, F.~Pirotte and M.~Saerens.
\newblock Random-walk computation of similarities between nodes of a graph,
  with application to collaborative recommendation.
\newblock In {\em IEEE Transactions on Knowledge and Data Engineering}, 19(3).

\bibitem{Franco2009silhouettes}
J.~S.~Franco and E.~Boyer.
\newblock Efficient Polyhedral Modeling from Silhouettes.
\newblock in {\em IEEE Transactions on Pattern Analysis and Machine Intelligence}, 31(3):414-427, 2009.

\bibitem{Franco2011learning}
J.~S.~Franco and E.~Boyer.
\newblock Learning Temporally Consistent Rigidities.
\newblock In {\em Computer Vision and Pattern Recognition}, 2011.

\bibitem{Furukawa05siggraph}
Y.~Furukawa and J.~Ponce.
\newblock Carved visual hulls for high accuracy image-based modeling.
\newblock In {\em Technical Sketch at SIGGRAPH}, 2005.

\bibitem{Golovinskiy09siggraph}
A.~Golovinskiy and T.~Funkhouser.
\newblock Consistent segmentation of 3d models.
\newblock In {\em SIGGRAPH}, 2009.

\bibitem{grauman03a}
K.~Grauman, G.~Shakhnarovich and T.~Darrell.
\newblock Inferring 3D structure with a statistical image-based shape model.
\newblock In {\em International Conference on Computer Vision}, 641--648, 2003.

\bibitem{gupta09understanding}
A.~Gupta, P.~Srinivasan, J.~Shi and L.~S. Davis.
\newblock Understanding videos, constructing plots learning a visually grounded
  storyline model from annotated videos.
\newblock In {\em Computer Vision and Pattern Recognition}, 2012--2019, 2009.

\bibitem{hayashi72}
C.~Hayashi.
\newblock Two dimensional quantification based on the measure of dissimilarity
  among three elements.
\newblock In {\em Ann. I. Stat. Math.}, 24:251--257, 1972.

\bibitem{heiser97triadic}
W.~J. Heiser and M.~Bennani.
\newblock Triadic distance models: Axiomatization and least squares
  representation.
\newblock In {\em J. Math. Psy.}, 41:189--206, 1997.

\bibitem{Hernandez04silhouette}
C.~Hernandez and F.~Schmitt.
\newblock Silhouette and stereo fusion for 3d object modeling.
\newblock {\em Computer Vision and Image Understanding}, 96(3):367--–392, 2004.

\bibitem{jain06smi}
V.~Jain and H.~Zhang.
\newblock Robust 3d shape correspondence in the spectral domain.
\newblock In {\em Shape Modeling International}, 2006.

\bibitem{toth01geometric}
N. Jakobson and J. Toth.
\newblock Geometric properties of eigenfunctions.
\newblock In {\em Russian Mathematical Surveys}, 56(6):1085--1106, 2001.

\bibitem{jenkins04icml}
O.~Jenkins and M.~Mataric.
\newblock A spatio-temporal extension to isomap nonlinear dimension reduction.
\newblock In {\em International Conference on Machine Learning}, 2004.

\bibitem{Katz05mesh}
S. Katz, G. Leifman and A. Tal.
\newblock Mesh segmentation using feature point and core extraction.
\newblock In {\em The Visual Computer}, 21:649--658, 2005.

\bibitem{kim09pami}
T.~Kim and R.~Cipolla.
\newblock Canonical correlation analysis of video volume tensors for action
  categorization and detection.
\newblock In {\em IEEE Transactions on Pattern Analysis and Machine Intelligence}, 31(8):1415--1428, 2009.

\bibitem{Knossow2008tracking}
D.~Knossow, R.~Ronfard, and R.~Horaud.
\newblock Human motion tracking with a kinematic parameterization of extremal contours
\newblock In {\em International Journal of Computer Vision}, 79 (3):247-269, 2008.

\bibitem{Lafon06pami}
S.~Lafon and A.~B. Lee.
\newblock Diffusion maps and coarse-graining: A unified framework for
  dimensionality reduction, graph partitioning, and data set parameterization.
\newblock In {\em IEEE Transactions on Pattern Analysis and Machine Intelligence}, 28(9).

\bibitem{levi06smi}
B. Levi.
\newblock Laplace-{B}eltrami eigenfunctions: {T}owards and algorithm that understands geometry.
\newblock In {\em Shape Modeling International}, 2006.

\bibitem{Lien07approximate}
J.-M. Lien and N.~Amanto.
\newblock Approximate convex decomposition of polyhedra.
\newblock In {\em Proceedings of the ACM Symposium on Solid and Physical
  Modeling}, 121--131, 2007.

\bibitem{lin06a}
R.~Lin, C.-B. Liu, M.-H. Yang, N.~Ahuja and S.~Levinson.
\newblock {Learning nonlinear manifolds from time series}.
\newblock In {\em European Conference on Computer Vision}, 2006.

\bibitem{Liu04segmentation}
R.~Liu and H.~Zhang.
\newblock Segmentation of 3D meshes through spectral clustering.
\newblock In {\em Proceedings of Computer Graphics and Applications}, pages
  298–--305, 2004.

\bibitem{Luxburg07tutorial}
U.~Luxburg.
\newblock A tutorial on spectral clustering.
\newblock In {\em Statistics and Computing}, 17(4).

\bibitem{macqueen67kmeans}
J.~MacQueen.
\newblock Some methods for classification and analysis of multivariate
  observations.
\newblock In {\em Proceedings of the Berkeley Symp. on Math. Stat. and Probability},
  volume~1, pages 281--297, 1967.

\bibitem{Mateus08cvpr}
D. Mateus, R. Horaud, D. Knossow, F. Cuzzolin and E. Boyer.
\newblock Articulated shape matching using laplacian eigenfunctions and
  unsupervised point registration.
\newblock In {\em Computer Vision and Pattern Recognition}, 2008.

\bibitem{Meila01aistats}
M.~Meila and J.~Shi.
\newblock A random walks view of spectral segmentation.
\newblock In {\em Artificial Intelligence and Statistics}, 2001.

\bibitem{Ng02nips}
A.~Ng, M.~Jordan and Y.~Weiss.
\newblock On spectral clustering: analysis and an algorithm.
\newblock In {\em Neural Information Processing Systems}, 2002.

\bibitem{Huang09shape}
Q.-X. Huang, M. Wicke, B. Adams and L. Guibas.
\newblock Shape decomposition using modal analysis.
\newblock In {\em EUROGRAPHICS}, volume~28, 2009.

\bibitem{Pons07multi}
J.-P. Pons, R. Keriven and O. Faugeras.
\newblock Multi-view stereo reconstruction and scene flow estimation with a
  global image-based matching score.
\newblock In {\em International Journal of Computer Vision}, 72(2):179--193,
  2007.

\bibitem{ralaivola04dynamical}
L.~Ralaivola and F.~d'Alche Buc.
\newblock Dynamical modeling with kernels for nonlinear time series prediction.
\newblock In {\em Neural Information Processing Systems}, volume~16, pages 129--136, 2004.

\bibitem{Roweis00}
S.~Roweis and L.~Saul.
\newblock Nonlinear dimensionality reduction by locally linear embedding.
\newblock In {\em Science}, 290(5500):2323--2326, 2000.

\bibitem{Saerens04ecml}
M. Saerens, F. Fouss, L. Yen and P. Dupont.
\newblock The principal components analysis of a graph, and its relationships
  to spectral clustering.
\newblock In {\em European Conference on Machine Learning},
  371--383, 2004.

\bibitem{Shamir08survey}
A.~Shamir.
\newblock A survey on mesh segmentation techniques.
\newblock In {\em Computer Graphics Forum}, 26:1539--1556, 2008.

\bibitem{Shapira08consistent}
L. Shapira, A. Shamir and D.~Cohen-Or.
\newblock Consistent mesh partitioning and skeletonisation using the shape
  diameter function.
\newblock In {\em The Visual Computer}, 24:249--–259, 2008.

\bibitem{sharma}
A. Sharma, R. Horaud, D. Knossow and E. von Lavante.
\newblock Mesh segmentation using Laplacian eigenvectors and Gaussian mixtures.
\newblock In {\em Proceedings of the AAAI Fall Symposium on Manifold Learning and its Applications}, 2009.

\bibitem{shi00normalized}
J.~Shi and J.~Malik.
\newblock Normalized cuts and image segmentation.
\newblock In {\em IEEE Transactions on Pattern Analysis and Machine Intelligence}, 22(8):888--905, 2000.

\bibitem{bb69551}
Q.~Shi, L.~Wang, L.~Cheng and A.~Smola.
\newblock Discriminative human action segmentation and recognition using
  semi-{M}arkov model.
\newblock In {\em Computer Vision and Pattern Recognition}, 2008.

\bibitem{Starck07correspondence}
J.~Starck and A.~Hilton.
\newblock Correspondence labeling for widetimeframe free-form surface matching.
\newblock In {\em International Conference on Computer Vision}, 2007.

\bibitem{sundaresan06a}
A.~Sundaresan and R.~Chellappa.
\newblock {Segmentation and probalistic registration of articulated body models}.
\newblock In {\em International Conference on Pattern Recognition}, 2006.

\bibitem{Tenenbaum00isomap}
J. B. Tenenbaum, V. de Silva and J. C. Langford.
\newblock A global geometric framework for nonlinear dimensionality reduction.
\newblock In {\em Science}, 290(5500):2319--2323, 2000.

\bibitem{varanasi}
K. Varanasi and E. Boyer.
\newblock Temporally coherent segmentation of 3D reconstructions.
\newblock In {\em Proceedings of 3DPTV}, 2010.

\bibitem{Zaharescu2011topology}
A.~Zaharescu, E.~Boyer, and R.~Horaud.
\newblock Topology-adaptive mesh deformation for surface evolution, morphing, and multi-view reconstruction.
\newblock In {\em IEEE Transactions on Pattern Analysis and Machine Intelligence}, 33(4):823-837, 2011.

\bibitem{vgl}
K. Zhou, J. Huang, J. Snyder, X. Liu, H. Bao, B. Guo and H.-Y. Shum.
\newblock Large mesh deformation using the volumetric graph Laplacian.
\newblock In {\em ACM Trans. Graph.}, 24(3):496--503, 2005.

\end{thebibliography}


\end{document}